\documentclass{article}
\usepackage{JFR_template}

\usepackage[T1]{fontenc}
\usepackage[latin9]{luainputenc}

\usepackage{amsmath}
\usepackage{amssymb}
\usepackage{graphicx}
\makeatletter
\usepackage{lmodern}

\usepackage[T1]{fontenc}
\usepackage{courier}
\usepackage{color}
\usepackage{upquote}
\usepackage{xcolor}
\usepackage{listings}
\usepackage{pgfplots}
\pgfplotsset{compat=1.16}
\usepackage{bm}
\usepackage{mathtools}
\usepackage{caption}
\usepackage{subcaption}
\usepackage{graphicx, epstopdf}
\usepackage{dsfont}
\usepackage{paralist}
\usepackage[nolist]{acronym}
\usepackage{algorithm,algorithmic}

\definecolor{hellgelb}{rgb}{1,1,0.85} 
\definecolor{colKeys}{rgb}{0,0,1} 
\definecolor{colIdentifier}{rgb}{0,0,0} 
\definecolor{colComments}{rgb}{0,0.5,0} 
\definecolor{colString}{rgb}{0.81,0.12,0.95}
\newcommand{\Tr}{\mathrm{T}} 

\DeclareMathOperator*{\argmin}{arg\,min}

\usepackage{xcolor}

\newcommand{\pr}[1]{\textbf{#1:}}  
\newcommand{\ph}[1]{\pr{#1}} 
\newcommand{\algorithmicbreak}{\textbf{break}}
\newcommand{\BREAK}{\STATE \algorithmicbreak}

\usepackage{microtype}

\usepackage{setspace}

\usepackage[subtle]{savetrees}

\usepackage{biblatex}
\usepackage{hyperref}

\hypersetup{
    colorlinks=false,
    linkcolor=blue,
    filecolor=magenta,      
    urlcolor=cyan,
}

\title{\LARGE \bf
STEP: Stochastic Traversability Evaluation and Planning for Risk-Aware Off-road Navigation; Results from the DARPA Subterranean Challenge
}

\author{
\parbox{0.6\linewidth}{\centering Anushri Dixit$^{*1}$, David D. Fan$^{*2}$, Kyohei Otsu$^{2}$, Sharmita Dey$^{2}$, Ali-Akbar Agha-Mohammadi$^{2}$, and Joel Burdick$^{1}$}
  \\\\
  \textsuperscript{1} California Institute of Technology\\
\textsuperscript{2} NASA Jet Propulsion Laboratory, California Institute of Technology\\
\textsuperscript{*} Indicates equal contribution. \\
{\textcopyright 2023, California Institute of Technology. All Rights Reserved}}%

\begin{document}
\maketitle
\begin{abstract}
Although autonomy has gained widespread usage in structured and controlled environments, robotic autonomy in unknown and off-road terrain remains a difficult problem.  Extreme, off-road, and unstructured environments such as undeveloped wilderness, caves, rubble, and other post-disaster sites pose unique and challenging problems for autonomous navigation. Based on our participation in the DARPA Subterranean Challenge, we propose an approach to improve autonomous traversal of robots in subterranean environments that are perceptually degraded and completely unknown through a traversability and planning framework called STEP (Stochastic Traversability Evaluation and Planning). We present 1) rapid uncertainty-aware mapping and traversability evaluation, 2) tail risk assessment using the Conditional Value-at-Risk (CVaR), 3) efficient risk and constraint-aware kinodynamic motion planning using sequential quadratic programming-based (SQP) model predictive control (MPC), 4) fast recovery behaviors to account for unexpected scenarios that may cause failure, and 5) risk-based gait adaptation for quadrupedal robots. We illustrate and validate extensive results from our experiments on wheeled and legged robotic platforms in field studies at the Valentine Cave, CA (cave environment),  Kentucky Underground, KY (mine environment), and Louisville Mega Cavern, KY (final competition site for the DARPA Subterranean Challenge with tunnel, urban, and cave environments).
\end{abstract}


\section{Introduction}

\begin{figure}[t!]
    \centering
    \includegraphics[width=\linewidth]{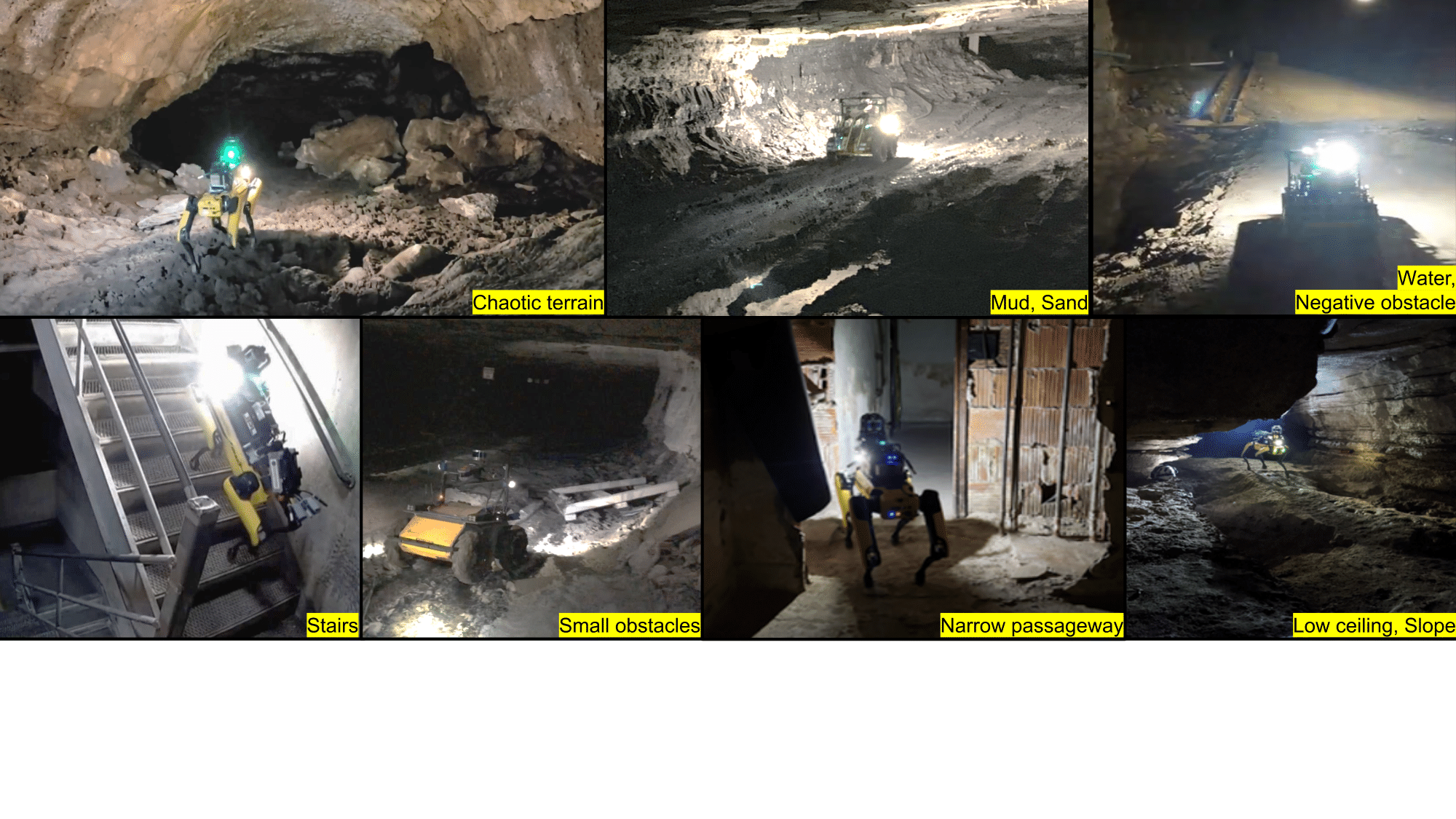}
    \vspace{-20mm}
    \caption{Top left: Boston Dynamics Spot quadruped robot exploring Valentine Cave at Lava Beds National Monument, CA.  Top middle, bottom middle (second image from the left):  Clearpath Husky robot exploring Arch Mine in Beckley, WV.  Bottom left:  Spot exploring abandoned Satsop power plant in Elma, WA.}
    \label{fig:HuskyArch}
\end{figure}
Robust autonomous traversal over extreme, hazardous terrain is an open problem with many applications to extra-terrestrial~\cite{ono2015chance}, disaster-struck~\cite{AliNeBula21}, and subterranean environments~\cite{kalita2018path}. The robots operate in uneven and highly risky terrain based on the use of noisy sensor measurements and localization uncertainty. Autonomous motion planning in such conditions requires a framework that can account for the traversability risk arising from sources like rubble, sudden drops, muddy/slippery areas, while also considering the uncertainty in the robots estimates. The framework must be able to make tractable reformulations of the nonconvex constraints arising from these risks so that it can plan reactive motions in real time. 

In this work, our evaluation of a terrain's \textit{traversability} accounts for the hazards from the uneven terrain (geometric, confidence-aware analysis) and type of terrain (semantic analysis), see~\autoref{fig:HuskyArch}. We will treat this traversability cost as a random variable and account for the uncertainty in the traversability cost using Conditional Value-at-Risk (CVaR). Using the CVaR assessment of the terrain, we obtain a geometric path and a low-level model predictive control (MPC) plan that accounts for the risk in the terrain.

Most traversability analyses are dependent on the types of sensors used to obtain information of the terrain, and are based on geometry, appearance, or proprioceptive methods \cite{papdakis2013survey}. Geometry-based methods often rely on building a 2.5D terrain map which is used to extract features such as maximum, minimum, and variance of the terrain height and slope~\cite{gestalt}. Planning algorithms for such methods take into account the stability of the robot on the terrain \cite{hait2002algorithms}. In \cite{otsu2020fast,ghosh2018pace}, the authors estimate the probability distributions of states based on the kinematic model of the vehicle and the terrain height uncertainty.  Furthermore, a method for incorporating sensor and state uncertainty to obtain a probabilistic terrain estimate in the form of a grid-based elevation map was considered in \cite{fankhauser2018probabilisticterrain}. Learning-based traversability frameworks are becoming increasingly popular. In~\cite{gan2022energy}, the authors proposed an inverse reinforcement learning-based method for learning the traversability cost from raw exterioceptive and proprioceptive sensor measurements. In~\cite{cai2022risk}, the authors learn a speed-map representation given semantic information while in~\cite{manderson2020learning}, the authors use a self-supervised learning-based terrain roughness classifier. In~\cite{fan2022learningSTEP}, the authors proposed a neural network architecture for learning the distribution of traversability costs arising from geometric risks using the pointcloud obtained from a LiDAR in a supervised manner. In this paper, we build upon the traversability analyses of classical geometric methods, while using semantic information from a LiDAR and incorporating the uncertainty of these methods for risk-aware planning \cite{agha2017CRM, thakker2021autonomous, fan2021step}.

Risk can be incorporated into motion planning using a variety of different methods, including chance constraints \cite{ono2015chance,wang2020non}, exponential utility functions \cite{koenig1994risk}, distributional robustness \cite{xu2010distributionally, wei2022moving}, and quantile regression \cite{fan2020deep,dabney2018distributional}. Risk measures, often used in finance and operations research, provide a mapping from a random variable (usually the cost) to a real number.  These risk metrics should satisfy certain axioms in order to be well-defined as well as to enable practical use in robotic applications \cite{majumdar2020should}.  Conditional value-at-risk (CVaR) is one such risk measure that has this desirable set of properties, and is a part of a class of risk metrics known as \textit{coherent risk measures} \cite{artzner1999coherent}.
Coherent risk measures have been used in a variety of decision making problems, especially Markov decision processes (MDPs) \cite{chow2015risk, ahmadi2020uncertainty, ahmadi2020constrained, ahmadi2021riskaverse}.  
Coherent risk measures have been used in a MPC framework when the system model is uncertain \cite{singh2018framework, dixit2022distributionally}, when the uncertainty is a result of measurement noise or moving obstacles \cite{hakobyan2019risk, dixit2020risksensitive}, and when the uncertainty arises from both measurement and process noise~\cite{dixit2022riskaverse}.  
  Our work extends CVaR risk to a risk-based planning framework which utilizes different sources of traversability risk (such as collision risk, step risk, slippage risk, etc.) Morever, this paper introduces the first field-hardened and theoretically grounded approach to traversability assessment and risk-constrained planning using CVaR metrics. Using CVaR to assess traversability risks allows us to dynamically tune the entire system's behavior - from aggressive to highly conservative - by changing a single value, the risk level. When the value of the risk level is small, the behavior of the traversability assessment is equivalent to an expectation-based assessment, which is similar to the geometric traversability evaluation in the past literature. Conversely, when the risk level is high, we account for the worst possible value of our terrain assessment given the uncertainty in our measurements. 

Model Predictive Control has long been used to robustly control more complex systems, including time-varying, nonlinear, or MIMO systems \cite{camacho2013model}.  While simple linear PID controllers are sufficient for simpler systems, MPC is well-suited to more complex tasks while being computationally feasible - MPC can account for the nonlinear robot dynamics while making multi-step predictions of the trajectory followed by the robot, and incorporating complex state and control constraints such as CVaR constraints, obstacle avoidance constraints, and non-holonomic constraints. Hence, MPC provides an optimization-based framework to mimic a suboptimal infinite-horizon controller while being computationally tractable.
In particular, we employ Sequential Quadratic Programming (SQP), which iteratively solves locally quadratic sub-problems to converge to a globally (more) optimal solution \cite{boggs1995sequential}.  This approach reduces computational costs and flexibility for handling a wide variety of costs and constraints \cite{schulman2014motion, lew2020chance, SCP1, scp2, nonlinearSCP}.  A common criticism of SQP-based MPC methods (and nonlinear MPC methods in general) is that they are susceptible to local minima.  We address this problem by incorporating a trajectory library (which can be predefined and/or randomly generated, e.g. as in \cite{kalakrishnan2011stomp}) to use in a preliminary trajectory selection process.  This approach finds more globally optimal initial guesses for the SQP problem to refine locally.  Another common difficulty with risk-constrained nonlinear MPC problems is ensuring recursive feasibility \cite{lofberg2012oops}.  We bystep this problem by dynamically relaxing the severity of the risk constraints while penalizing CVaR in the cost function. 

In this work, we propose STEP (Stochastic Traversability Evaluation and Planning), that pushes the boundaries of the state-of-the-practice to enable safe, risk-aware, and high-speed ground traversal of unknown environments.  Specifically, our contributions include:
\begin{enumerate}
    \item Uncertainty-aware 2.5D traversability evaluation which accounts for localization error, sensor noise, and occlusion, and combines multiple sources of traversability risk.
    \item An approach for combining these traversability risks into a unified risk-aware CVaR planning framework.
    \item A highly efficient MPC architecture for robustly solving non-convex risk-constrained optimal control problems.
    \item A suite of recovery behaviors to account for fast response to failure scenarios.
    \item Risk-based gait adaptation for quadrupedal robots (in our case, the Boston Dynamics Spot platform).
    \item Real-world demonstration of real-time CVaR planning on wheeled and legged robotic platforms in unknown and risky environments.
\end{enumerate}
This work significantly extends our previous work~\cite{fan2021step} to incorporate semantic and confidence-aware traversability risk sources in addition to geometric ones to provide a complete suite of traversability analyses using LiDARs. Further, we provide recovery behaviors in our hierarchical planning framework to account for high-risk situations that the geometric and kinodynamic planners fail to account for. Lastly, we also provide risk-aware gait selection techniques for quadrupedal robots. This hierarchical framework is tested extensively in multiple subterranean environments.



\section{Risk-Aware Traversability and Planning}
\subsection{Problem Statement}


%
Let $x_k$, $u_k$, $z_k$ denote the robot's state, action (or control input), and observation (or sensory measurement) at the $k$-th time step. A path $x_{0:N}=\{x_0, x_1, \cdots, x_N\}$ is composed of a sequence of poses. A policy is a mapping from state to control $u_k = \pi(x_k)$. A map is represented as $m = (m^{(1)}, m^{(2)}, \cdots)$ where $m^{i}$ is the $i$-th element of the map (e.g., a cell in a grid map or feature in a feature-based map).
%
The robot's dynamics model captures the physical properties of the vehicle's motion, such as inertia, mass, dimension, shape, and kinematic and control constraints:
\begin{align}
    x_{k+1} &= f(x_{k}, u_{k}) \\
    g(u_{k}) &\succ 0 \\
    h(m, x_{k}) &\succ 0
\end{align}
where $g(u_k),$ $ h(m, x_{k})$ are vector-valued functions that encode the control and state constraints/limits respectively.

Following \cite{papdakis2013survey}, we define \textit{traversability} as the capability for a ground vehicle to reside over a terrain region under an admissible state that satisfies constraints.  We represent traversability as a cost, i.e. a continuous value computed using a terrain model, the robotic vehicle model, and kinematic constraints, which represents the degree to which we wish the robot to avoid a given state:
\begin{align}
    r = \mathcal{R}(m, x, u)
\end{align}
where $r\in\mathbb{R}$, and $\mathcal{R}(\cdot)$ is a terrain assessment model.  This model captures various unfavorable events such as collision, getting stuck, tipping over, high slippage, to name a few. Each mobility platform has its own assessment model to reflect its mobility capability.

Associated with the true traversability value is a distribution over possible values based on the current understanding about the environment and robot actions. In most real-world applications where perception capabilities are limited and noisy, the true value can be highly uncertain.  To handle this uncertainty, consider a map belief, i.e., a probability distribution $p(m | x_{0:k}, z_{0:k})$, over a possible set $\mathcal{M}$.  Then, the traversability estimate is also represented as a random variable $R: (\mathcal{M} \times \mathcal{X} \times \mathcal{U}) \longrightarrow \mathbb{R}$.  We call this probabilistic mapping from map belief, state, and controls to possible traversability cost values a \textit{terrain assessment model}.

A risk metric $\rho(R):R\rightarrow\mathbb{R}$ is a mapping from a random variable to a real number which quantifies some notion of risk.  In order to assess the risk of traversing along a path $x_{0:N}$ with a policy $\pi$, we wish to define the cumulative risk metric associated with the path, $J(x_0,\pi)$.  To do this, we need to evaluate a sequence of random variables $R_{0:N}$.  To quantify the stochastic outcome as a real number, we use the dynamic, time-consistent risk metric given by compounding the one-step risk metrics \cite{ruszczynski2014riskaverseDP}:
\begin{align}
  J(x_0, \pi; m)
  &= R_0 + \rho_0\big( R_1 + \rho_{1}\big(R_2 + \dotsc + \rho_{N-1}\big(R_{N})\big)\big)
  \label{eq:risk_metric}
\end{align}
where $\rho_k(\cdot)$ is a one-step coherent risk measure at time $k$. This one-step risk gives us the cost incurred at time-step $k+1$ from the perspective of time-step $k$. Any distortion risk measure compounded as given in \eqref{eq:risk_metric} is time-consistent (see \cite{majumdar2020should} for more information on distortion risk metrics and time-consistency). We use the Conditional Value-at-Risk (CVaR) as the one-step risk metric:
\begin{align}
    \rho(R) =
    \mathrm{CVaR}_{\alpha}(R)=
    &\inf_{z \in \mathbb{R}} \mathbb{E}\Bigg[z + \frac{(R-z)_{+}}{1-\alpha}\Bigg]
\end{align}
where $(\cdot)_{+}=\max(\cdot, 0)$, and $\alpha \in (0, 1]$ denotes the \textit{risk probability level}. We note that the results in this paper are also easily extended to other tail risk measures like Entropic Value-at-Risk~\cite{ahmadi2012entropic} and total variation distance-based risk~\cite{dixit2022tvd}.

We formulate the objective of the problem as follows: Given the initial robot configuration $x_{S}$ and the goal configuration $x_{G}$, find an optimal control policy $\pi^{*}$ that moves the robot from $x_{S}$ to $x_{G}$ while 1) minimizing time to traverse, 2) minimizing the cumulative risk metric along the path, and 3) satisfying all kinematic and dynamic constraints. For a quadrupedal robot like the Boston Dynamics Spot robot, the framework must additionally also select the best gait type based on the risk accrued while moving from $x_{S}$ to $x_{G}$ as a part of the optimal policy $\pi^{*}$.

\subsection{Hierarchical Risk-Aware Planning}



We propose a hierarchical approach to address the aforementioned risk-aware motion planning problem by splitting the motion planning problem into geometric and kinodynamic domains.  We consider the geometric domain over long horizons, while we solve the kinodynamic problem over a shorter horizon. 
This is convenient for several reasons:  1) Solving the full constrained CVaR minimization problem over long timescales/horizons becomes intractable in real-time.  2) Geometric constraints play a much larger role over long horizons, while kinodynamic constraints play a much larger role over short horizons (to ensure dynamic feasibility at each timestep).  3) A good estimate (upper bound) of risk can be obtained by considering position information only.  This is done by constructing a position-based traversability model $\mathcal{R}_{\mathrm{pos}}$ by marginalizing out non-position related variables from the terrain assessment model, i.e. if the state $x=[p_x,p_y,x_{\mathrm{other}}]^{\intercal}$ consists of position and non-position variables (e.g. orientation, velocity), then
\begin{align}
        \mathcal{R}_{\mathrm{pos}}(m,p_x,p_y) \geq \mathcal{R}(m,x,u) \quad \forall x_{\mathrm{other}}, u
\end{align}


\textit{Geometric Planning:} The objective of geometric planning is to search for an \textit{optimistic} risk-minimizing path, i.e. a path that minimizes an upper bound approximation of the true CVaR value.  For efficiency, we limit the search space only to the geometric domain.  We search for a sequence of poses $x_{0:N}$ which ends at $x_G$ and minimizes the position-only risk metric in \eqref{eq:risk_metric}, which we define as $J_{\mathrm{pos}}(x_{0:N})$. The path optimization problem can be written as:
\begin{align}
    x_{0:N}^{*} = \argmin_{x_{0:N}} \bigg[ J_{\mathrm{pos}}(x_{0:N}) &+ \lambda \sum_{k=0}^{N-1} \|x_k - x_{k+1}\|^2 \bigg]\label{eq:optimization_geometric_cost}\\
    s.t. \quad \phi(m, x_k) &\succ 0
    \label{eq:optimization_geometric_constraint}
\end{align}
where the constraints $\phi(\cdot)$ encode position-dependent traversability constraints (e.g. constraining the vehicle to avoid obstacles and prohibit lethal levels of risk) and $\lambda\in\mathbb{R}$ weighs the tradeoff between risk and path length.




\begin{figure*}[t!]
\vspace{-10mm}
    \centering
    \includegraphics[width=0.99\textwidth]{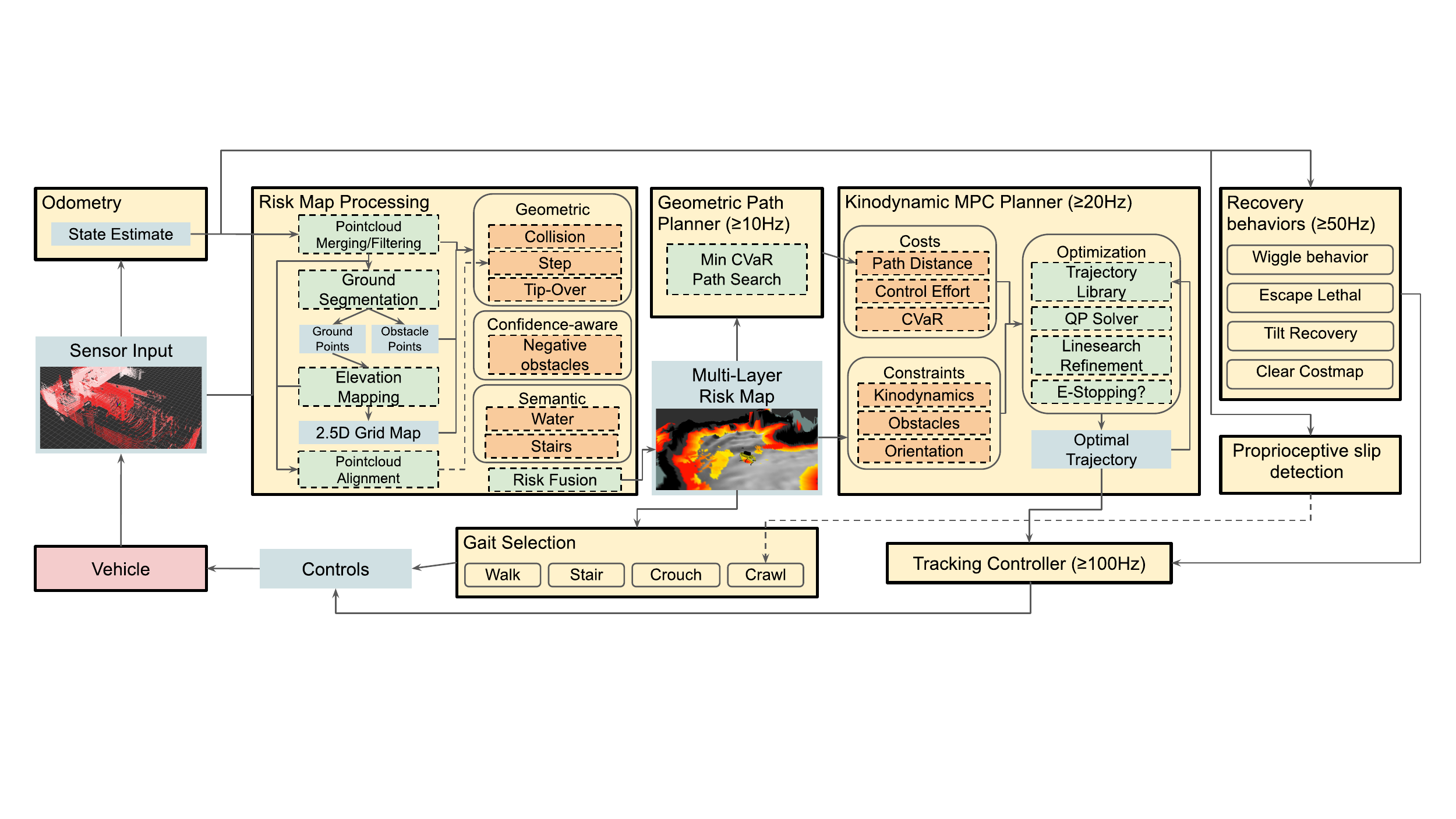}
    \vspace{-25mm}
    \caption{Overview of system architecture for STEP.  From left to right:  Odometry aggregates sensor inputs and relative poses.  Next, Risk Map Processing merges these pointclouds and creates a multi-layer risk map.  The map is used by the Geometric Path Planner and the Kinodynamic MPC Planner.  An optimal trajectory is found and sent to the Tracking Controller, which produces control inputs to the robot.}
    \label{fig:system}
\end{figure*}

\textit{Kinodynamic Planning:} We then solve a kinodynamic planning problem to track the optimal geometric path, minimize the risk metric, and respect kinematic and dynamics constraints. The goal is to find a control policy $\pi^{*}$ within a local planning horizon $T\leq N$ which tracks the path $X^*_{0:N}$.  The optimal policy can be obtained by solving the following optimization problem:

\begin{align}
 \pi^{*} = \argmin_{\pi \in \Pi} \bigg[
    J(x_0, \pi) + &\lambda \sum_{k=0}^T \|x_{k} - x^*_{k}\|^2
    \bigg]
    \label{eq:mpc_cost}\\
 s.t. ~ \forall k\in[0,\cdots,T]: \qquad x_{k+1} &= f(x_{k}, u_{k}) \label{eq:mpc_dynamics}\\ 
    g(u_k) &\succ 0 \label{eq:mpc_control_constraint}\\ 
    h(m, x_k) &\succ 0 
    \label{eq:mpc_state_constraints}
\end{align}
where the constraints $g(u)$ and $h(m, x_k)$ are vector-valued functions which encode controller limits and state constraints, respectively.

\textit{Recovery from Unfavorable States:}
Recovery planning is a particular type of planning problem where the initial state is not safe. For example, the robot might need to start planning from the state where it touches walls with its bumpers, or the state where the body is tilted on top of rubble. The recovery from those unfavorable states involves finding a safe control without violating the safety constraints under smaller margin conditions.

\section{STEP for Unstructured Terrain}

This section discusses how we compute traversability risk and efficiently solve the risk-aware trajectory optimization problem.  At a high level, our approach takes the following steps (see Figure \ref{fig:system}):  1) Aggregate sensor measurements to create an uncertainty-aware map using some source of localization with uncertainty,  2)  Perform ground segmentation to isolate the parts of the map that the robot can potentially traverse.  3)  Compute risk and risk uncertainty using geometric properties of the pointcloud (optionally, include other sources of risk, e.g. semantic or other sensors).  4) Aggregate these risks to compute a 2.5D CVaR risk map.  5)  Solve for an optimistic CVaR minimizing path over long ranges with a geometric path planner.  7) Solve for a kinodynamically feasible trajectory which minimizes CVaR while staying close to the geometric path and satisfying all constraints.

\subsection{Pointcloud Processing and Mapping}

\ph{Multi-sensor Merging}
Our pointcloud pipeline starts from merging pointclouds from different sensors. One robot can have multiple units of the same senor to increase coverage, or have heterogeneous sensors that produce pointclouds using different mechanics (e.g., active LiDAR, RGBD cameras, passive stereo cameras). After applying sensor-specific filters that remove noise or body occlusion, these pointclouds are merged using extrinsic calibration information. If the sensors are not time-synchronized, we use odometry to compensate motion offset.

\ph{Temporal Pointcloud Merging}
The merged pointclouds are aggregated over a fixed time window to construct a local pointcloud map. We maintain a pose graph on the time window to incorporate history updates in the odometry estimate (e.g., loop closures). Based on the latest pose graph, we reconstruct a full aggregated pointcloud. We annotate each point by the time offset to the latest pointcloud. This allows us to propagate odometry uncertainty to each point in the fusion phase.

\ph{Ground Segmentation}
The aggregated pointcloud is segmented into ground and obstacle points using 3D pointcloud segmentation techniques. We leverage the work in~\cite{himmelsbach2010fast} that allows efficient ground segmentation based on line fitting in the cylindrical coordinates. We extended the work to also handle challenges prevalent in subterranean environments, such as low ceiling or negative obstacles. The ground segmentation is critical for elevation mapping in an occluded environment where the ground is not observed by the sensors and the measurements of walls/ceiling make false ground planes.

\begin{figure}
    \centering
    \includegraphics[width=0.6\textwidth]{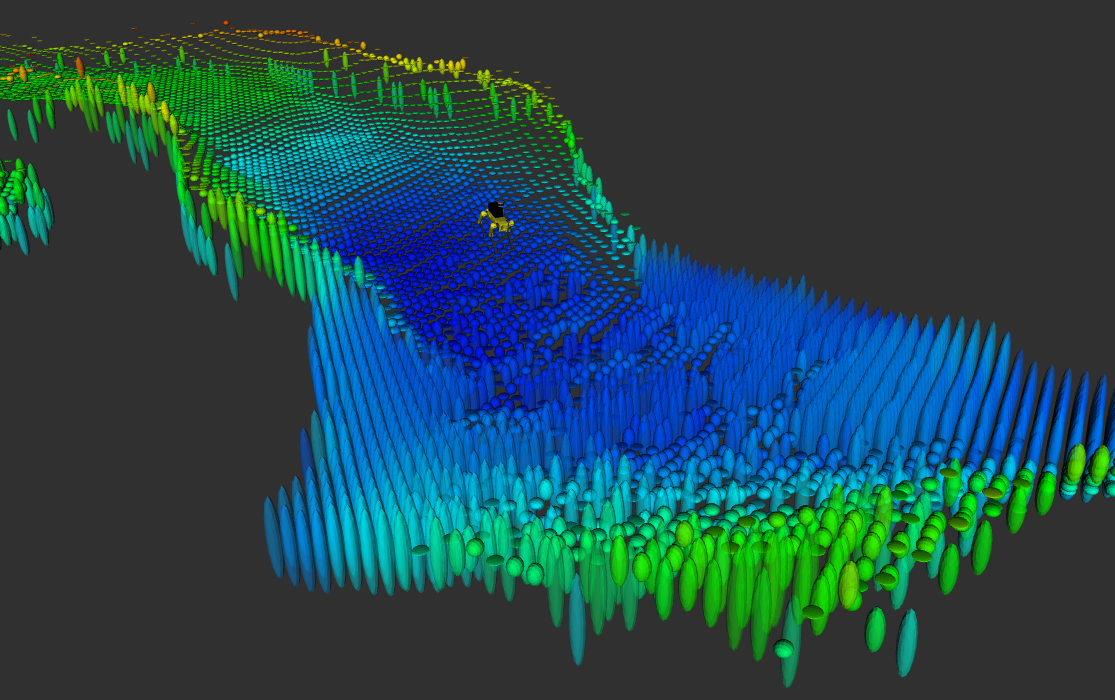}
    \caption{Estimation of ground height with uncertainty}
    \label{fig:belief_cloud}
\end{figure}

\ph{Elevation Mapping}
We construct 2.5D height map using the segmented ground points. After splitting points to each grid map cell and sorting by time stamps, we apply a Kalman filter to estimate the height of the ground. We set up the filter to place more weight on the recent measurements which are less affected by the odometry error. \autoref{fig:belief_cloud} shows the visualization of ground height estimation with uncertainty. Note that there is higher level of uncertainty for the areas that have few measurements. These mapping uncertainties are used to adjust confidence in a later traversability estimation process.

\subsection{Modeling Assumptions}

Among many representation options for rough terrain, we use a 2.5D grid map in this paper for its efficiency in processing and data storage \cite{Fankhauser2016GridMapLibrary}. The map is represented as a collection of terrain properties (e.g., height, risk) over a uniform grid.

For different vehicles we use different robot dynamics models.  For example, for a system which produces longitudinal/lateral velocity and steering (e.g. legged platforms), the state and controls can be specified as:
\begin{align}
    x &= [p_x, p_y, p_\theta, v_x, v_y, v_{\theta}]^{\intercal} \\
    u &= [a_x, a_y, a_{\theta}]^{\intercal}
\end{align}
While the dynamics $x_{k+1} = f(x_k,u_k)$ can be written as,
\begin{align*}
    x_{k+1} = x_k + \Delta t \begin{bmatrix}
        v_x \cos(p_\theta) - v_y \sin(p_\theta) \\
        v_x \sin(p_\theta) + v_y \cos(p_\theta)\\
        \kappa v_x + (1-\kappa) v_{\theta}\\
        a_x\\
        a_y\\
        a_{\theta}
    \end{bmatrix}.
\end{align*}
We let $\kappa\in[0,1]$ be a constant which adjusts the amount of turning-in-place the vehicle is permitted.  In differential drive or ackermann steered vehicles we can remove the lateral velocity component of these dynamics.  However, our general approach is applicable to any vehicle dynamics model.  (For a differential drive model, see Appendix~\ref{appendix:dyn})





\subsection{Terrain assessment models}

\begin{figure}[th!]
    \centering
    \includegraphics[width=0.9\linewidth]{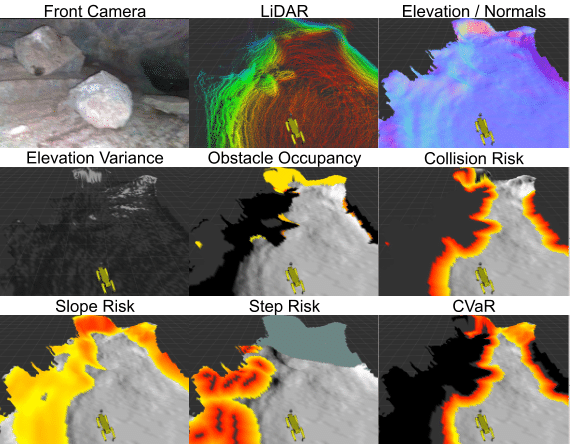}
    \caption{Multi-layer geometric risk analysis, which first aggregates recent pointclouds (top).  Then, each type of analysis (slope, step, collision, etc.) generates a risk map along with uncertainties (middle rows).  These risks are aggregated to compute the final CVaR map (bottom).}
    \label{fig:trav_challenges}
\end{figure}

 The traversability cost is assessed as the combination of multiple risk factors. These factors are designed to capture potential hazards for the target robot in the specific environment (Figure \ref{fig:trav_challenges}). Such factors include:
\begin{itemize}
    \item \textit{Collision}: quantified by the distance to the closest obstacle point.
    \item \textit{Step size}: the height gap between adjacent cells in the grid map.
    \item \textit{Tip-over}: a function of slope angles and the robot's orientation.
    \item \textit{Sensor Uncertainty}:  sensor and localization error increase the variance of traversability estimates.
    \item \textit{Negative Obstacles}: detected by checking the lack of measurement points in a cell.
    \item \textit{Slippage}: quantified by geometry and the surface material of the ground. 
\end{itemize}

\subsubsection{Geometry-based risk sources}
Geometry-based risk sources include collision with obstacles, too-large step sizes in terrain, and impassable slopes.  These geometry-based risks are constructed using geometric analysis of elevation map estimates and LiDAR pointcloud points.  We construct these risks per grid cell, with the following methods.  First, using the ground estimates from the ground segmentation pipeline, we obtain a lower bound on the height of the ground, i.e. the height at which the robot would place its foot or wheel if occupying that cell with its foot/wheel.  Above this ground estimate, we can determine the relative height of other occupying LiDAR points in the temporally merged pointcloud with respect to the ground.  Points which occur above or below a certain height threshold which outline the body of the robot are treated as obstacles, and the corresponding grid cells at these points are marked as untraversable.  Similar analysis is performed for step size risk, which checks the height gap between adjacent cells in the elevation map.  For adjacent cells which exhibit too high a height gap, these cells are marked as untraversable.  Finally, checking the normals of the elevation map~\cite{Fankhauser2016GridMapLibrary} gives an overall estimation of slope (which the size of the normal calculation averaged over the approximate size of the robot),  and areas with too high slope values are marked as high risk or untraversable.

Note that in all these analyses, the uncertainty of the elevation map plays a large role. Sensor and localization uncertainty corrupts elevation measurements to a varying degree, proportional with distance from the robot.  Therefore we adapt various detection or risk thresholds with distance from the robot, to obtain a more robust result.  For example, for the Husky robot, ground clearance beneath the robot belly is 10cm.  Detecting a 10cm step size in elevation at 100m away requires an angular localization accuracy of < 0.06 degrees.  This may be infeasbile, and therefore adapting the step size threshold with distance helps to reduce spurious detections at longer ranges.  
\subsubsection{Confidence-based risk sources}
\begin{figure}[th!]
    \centering
    \includegraphics[width=\linewidth]{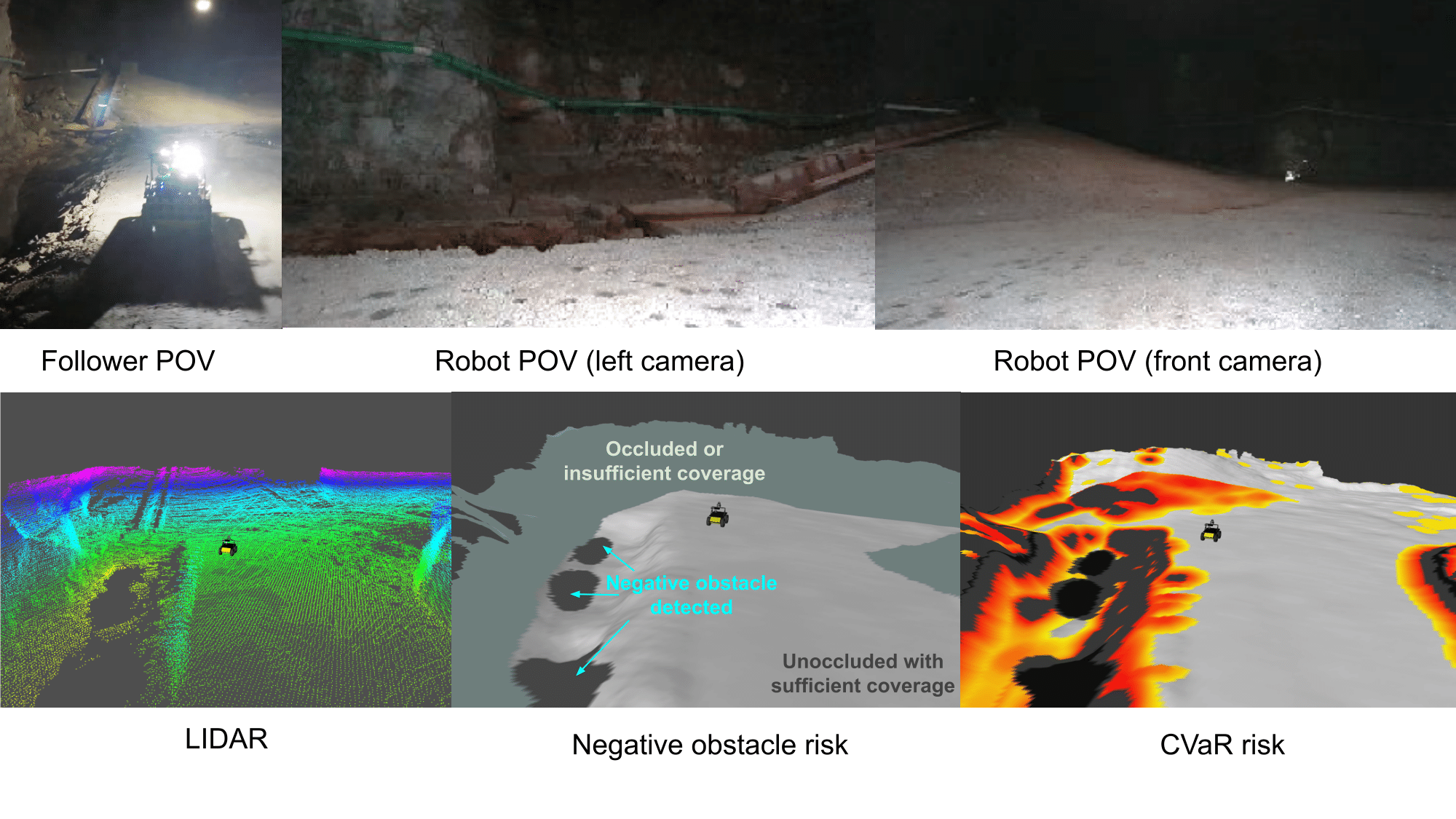}
    \vspace{-5mm}
    \caption{Confidence-based risk analysis: the scene is illustrated through the point-of-view (POV) of a third-person (top left) and robot (top middle and top right). The aggregated pointcloud (bottom left) has regions of no returns from the area on the left side of the robot. These holes in the pointcloud are marked as negative obstacles in the risk layer (bottom middle) only when there are no returns from these regions despite them being unoccluded and sufficiently covered by laser strike pattern. This risk layer is aggregated with the geometric risk layers to compute the final CVaR map (bottom right).}
    \label{fig:neg_obs_risk}
\end{figure}

One example of confidence-based risk sources is the detection of negative obstacles. To detect negative obstacles, we estimate areas in the pointcloud that have no returns and use 2-D ray-tracing to find "gap" areas that are not occluded by obstacles, small steps, or upward slopes. However, if this is the only criteria for negative obstacle detection, many false positives for negative obstacles are observed. One such instance is shown in Fig~\ref{fig:neg_obs_risk} when the robot turns the corner into a new room from a narrow passageway and in the time-taken for the LiDAR returns from all areas of the new room to reach the sensor, false-positives of negative obstacles are detected. To address this, we account for the confidence in the gaps in the pointcloud by estimating whether these areas have been sufficiently covered so far, i.e., the robot has sensed the area from different positions and for long enough to ensure that the gaps in the pointcloud are not caused due to the sensor and environment configuration, see Algorithm~\ref{alg:coverge_algorithm} for a detailed description of the method to check whether a given region passes the coverage check. 
\begin{algorithm}[t]
    \algsetup{linenosize=\small}
    \small
    \caption{Algorithm for computing whether a given area is unoccluded and sufficiently covered, i.e., $m^{\text{cover}}_{\text{MSE}} >= 0$ }
    \begin{algorithmic}[1]
        \renewcommand{\algorithmicrequire}{\textbf{Input:}}
        \renewcommand{\algorithmicensure}{\textbf{Output:}}
        \renewcommand{\algorithmiccomment}{\textbf{\#}}
        \REQUIRE  $m^{\text{unocc}}$: Grid map of unoccluded grid points from the current robot state $x_t$ (using ray-tracing), \\
        $x_t = [p_x, p_y,  p_\theta, v_x, v_y, v_\theta]^T$: Current robot state, \\
        $d^{\text{cover}}$: Threshold for sufficient coverage (in meters).
        \ENSURE $m_N^{\text{unocc}}$: Number of times each grid cell in the grid map has been observed unoccluded by the robot so far, \\$m^{\text{cover}} = \{m^{\text{cover}}_{\mu_x}, m^{\text{cover}}_{\mu_y}, m^{\text{cover}}_{\text{MSE}}\}$: Multi-layer map indicating whether each grid cell being evaluated for presence of negative obstacles has been observed unoccluded from different points to ensure sufficient coverage.
        \FOR {$i = 0$ to $|m^{\text{unocc}}|$} 
        \STATE { \COMMENT{ \textit{Update $m_N^{\text{unocc}}$}} \\
        $m_N^{\text{unocc},(i)} = \left\{ \begin{array}{ll}
                  1 +m_N^{\text{unocc},(i)} &\quad \text{if } m^{\text{unocc},(i)} = 1\text{ (i.e., $i^{\text{th}}$ grid cell is unoccluded)}\\
                  m_N^{\text{unocc},(i)} &\quad \text{if } m^{\text{unocc},(i)} = 0\text{ (i.e., $i^{\text{th}}$ grid cell is occluded)}
                \end{array}
              \right.$}
         \STATE {\COMMENT{ \textit{ Update mean $x$ and $y$ robot positions from where each grid cell has been observed unoccluded } \\}
        $\{m^{{\text{cover}}, (i)}_{\mu_x}, m^{{\text{cover}}, (i)}_{\mu_y}\} = \left\{ \begin{array}{ll}
                  \{m^{{\text{cover}}, (i)}_{\mu_x} + \frac{p_x - m^{{\text{cover}}, (i)}_{\mu_x}}{m_N^{\text{unocc},(i)}}, m^{{\text{cover}}, (i)}_{\mu_y}+ \frac{p_y - m^{{\text{cover}}, (i)}_{\mu_y}}{m_N^{\text{unocc},(i)}}\}  &\quad \text{if } m^{\text{unocc},(i)} = 1\\
                  \{m^{{\text{cover}}, (i)}_{\mu_x}, m^{{\text{cover}}, (i)}_{\mu_y}\}  &\quad \text{if } m^{\text{unocc},(i)} = 0
                \end{array}
              \right.$}
        \STATE{ \COMMENT{ \textit{ Update mean squared deviation of the average robot positions from which a given grid cell is observed \\ to check if it crosses the threshold for sufficient coverage, $d^{\text{cover}}$}} \\
     $m^{{\text{cover}}, (i)}_{\text{MSE}} = \left\{ \begin{array}{ll} (p_x-m^{{\text{cover}}, (i)}_{\mu_x})^2 + (p_y - m^{{\text{cover}}, (i)}_{\mu_y})^2 - (d^{\text{cover}})^2 &\quad \text{if } m^{\text{unocc},(i)} = 1\\
    - (d^{\text{cover}})^2  &\quad \text{if } m^{\text{unocc},(i)} = 0\end{array}
              \right.$}
    \ENDFOR
    \end{algorithmic} 
    \label{alg:coverge_algorithm}
\end{algorithm}

\subsubsection{Semantics-based risk sources}
\begin{figure}[th!]
    \centering
    \includegraphics[width=0.9\linewidth]{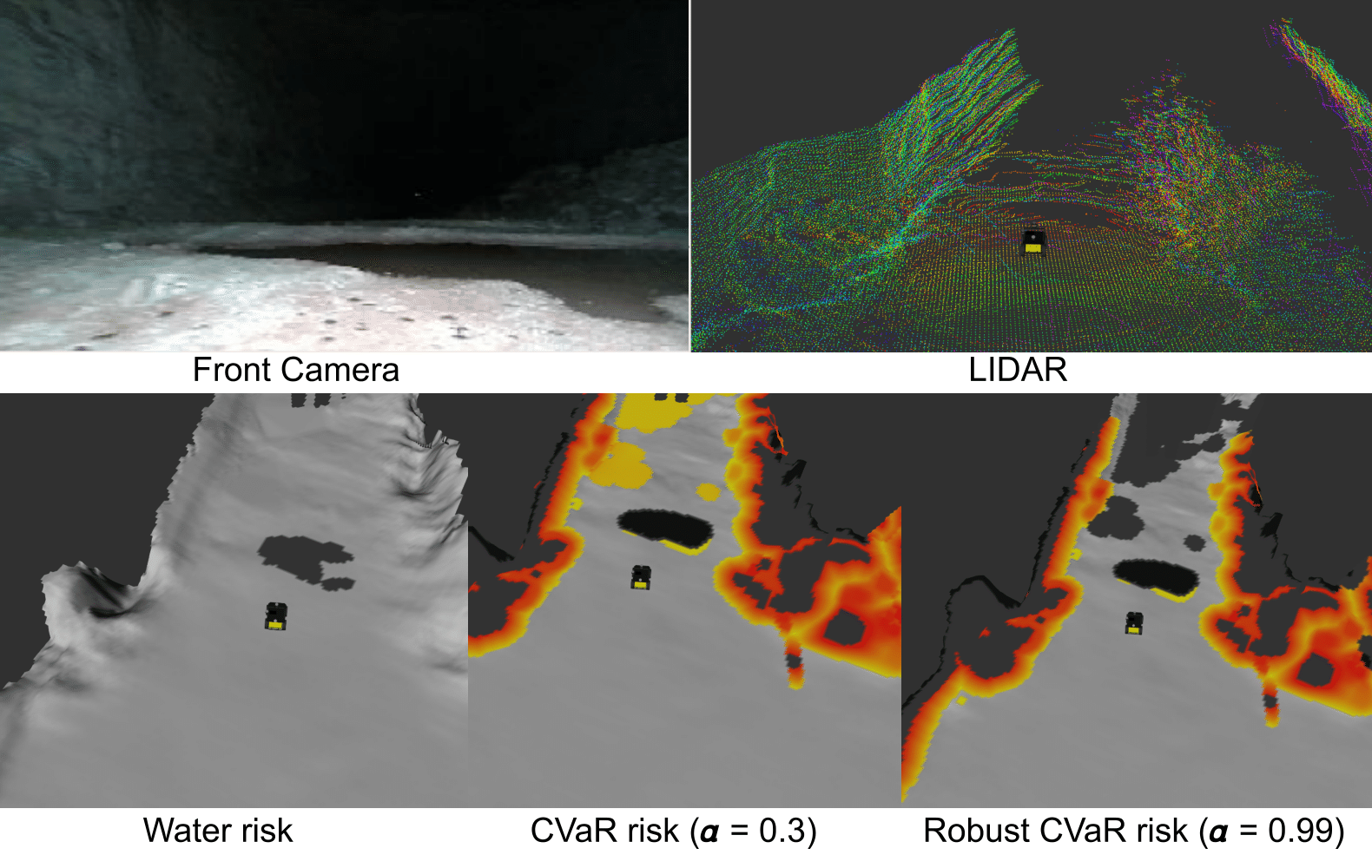}
    \caption{Semantics-based risk analysis: the robot camera (top left) shows a puddle of water in front of the robot. The aggregated pointcloud (top right) has regions of no returns from the area in front of the robot. These holes in the pointcloud are marked as water in the risk layer (bottom left) when the holes in the pointcloud are present with low intensity returns in the area near the hole. This risk layer is aggregated with the geometric risk layers to compute the final CVaR map (bottom middle and bottom right) with different levels of robustness $\alpha$. We clearly see the effects on the risk map, where higher values of $\alpha$ result assigning high cost to unknown regions.}
    \label{fig:water_risk}
\end{figure}

LiDAR intensity returns are different on different materials at the same distance from the sensor. In~\cite{li2008intensityClassifier}, the authors demonstrate the efficacy of terrain classification using LiDAR intensity and height returns. We account for changes in the terrain features based on changes in the intensity of LiDAR returns. Areas that comprise a lot of mud and water have low intensity of returns. This enables detection of muddy regions where the robot may get stuck or fall down. Areas with deep water levels are similarly detected as negative obstacles because the LiDAR returns are very sparse. Notably, negative obstacles combined with low intensity of returns allows us to detect lethal regions with deep water levels.~\autoref{fig:water_risk} shows an instance of the water-detection when the robot is near a water puddle. Note that in~\autoref{fig:neg_obs_risk}, on the contrary, while there are holes in the LiDAR pointcloud, they are not surrounded by low intensity returns and hence classified as negative obstacles.
\subsection{Traversability Cost}
To efficiently compute the Conditional Value-at-Risk (CVaR) traversability cost for $l>1$ risk factors, we assume each risk factor $R_l$ is an independent random variable which is normally distributed, with mean $\mu_l$ and variance $\sigma_l$.  We take a weighted average of the risk factors to obtain the terrain assessment model, $R=\sum_lw_lR_l$, which will also be normally distributed as $R\sim\mathcal{N}(\mu,\sigma^2)$. Let $\varphi$ and $\boldsymbol{\Phi}$ denote the probability density function and cumulative distribution function of a standard normal distribution respectively.  The corresponding CVaR is computed as:
\begin{align}
  \rho(R) = \mu + \sigma\frac{\varphi(\boldsymbol{\Phi}^{-1}(\alpha))}{1-\alpha}
  \label{eq:cvar_normal}
\end{align}
The CVaR cost accounts for the expected cost in the $(1-\alpha)$ tail of the distribution of the terrain assessment model, $R$, thus also accounting for high-risk, low probability events. We construct the $R$ such that the expectation of $R$ is positive, to keep the CVaR value positive, $\mu \geq 0$.

Construction of the mean and variance of each risk factor depends on the type of risk.  For example, collision risk is determined by checking for points above the elevation map, and the variance is derived from the elevation map variance, which is mainly a function of localization error.  In contrast, negative obstacle risk is determined by looking for gaps in sensor measurements.  These gaps tend to be a function of sensor sparsity, so the risk variance increases with distance from the sensor frame. 



\subsection{Risk-aware Geometric Planning}

In order to optimize \eqref{eq:optimization_geometric_cost} and \eqref{eq:optimization_geometric_constraint}, the geometric planner computes an optimal path that minimizes the position-dependent dynamic risk metric in \eqref{eq:risk_metric} along the path. Substituting \eqref{eq:cvar_normal} into \eqref{eq:risk_metric}, we obtain:
\begin{align}
    J_{\mathrm{pos}}(x_{0:N})
    &= \mu_0 + \sum_{k=1}^N \bigg[ \mu_k + \sigma_k \frac{\varphi(\boldsymbol{\Phi}^{-1}(\alpha))}{1-\alpha} \bigg]
\end{align}
(For a proof, see Appendix \ref{appendix:cvar}.)

We use the A$^*$ algorithm to solve \eqref{eq:optimization_geometric_cost} over a 2D grid. A$^*$ requires a path cost $g(n)$ and an admissible heuristic cost $h(n)$, given by:
\begin{align}
    g(n) &= J_{\mathrm{pos}}(x_{0:n}) + \lambda \sum_{k=0}^{n-1} \|x_k - x_{k+1}\|^2, \\
    h(n) &= \lambda\,\|x_{n} - x_{G}\|^2.
\end{align}
For the heuristic cost we use the shortest Euclidean distance to the goal, which is always less than the actual path cost. The parameter lambda is a relative weighting between the distance penalty and risk penalty and can be thought of as having units of (traversability cost / m).  Our experiments use a relatively small value, which means we are mainly concerned with minimizing traversability costs.  


\subsection{Risk-aware Kinodynamic Planning}
\begin{figure}[t!]
    \centering
    \includegraphics[width=150mm]{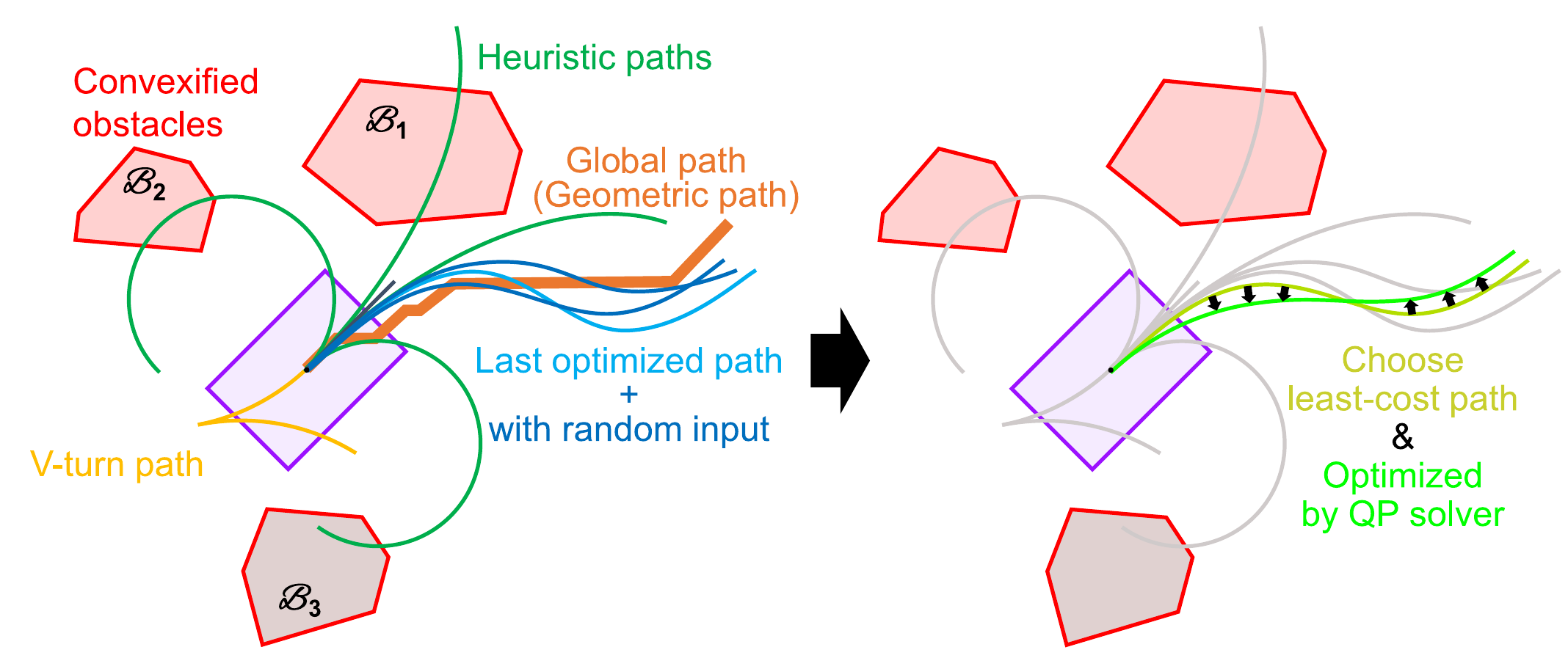}
    \caption{Diagram of kinodynamic MPC planner, which begins with evaluating various paths within a trajectory library.  The lowest cost path is chosen as a candidate and optimized by the QP solver.}
    \label{fig:traj_vis}
\end{figure}

The geometric planner produces a path, i.e. a sequence of poses.  We wish to find a kinodynamically feasible trajectory which stays near this path, while satisfying all constraints and minimizing the CVaR cost.  To solve \eqref{eq:mpc_cost}-\eqref{eq:mpc_state_constraints}, we use a risk-aware kinodynamic MPC planner, whose steps we outline (Figure \ref{fig:traj_vis}, Algorithm \ref{alg:full_algorithm}, Appendix \ref{appendix:traj_fig}).

\begin{algorithm}[t]
    \algsetup{linenosize=\small}
    \small
    \caption{Kinodynamic MPC Planner (sequences $\{\text{var}_k\}_{k=0:T}$ are expressed as $\{\text{var}\}$ for brevity)}
    \begin{algorithmic}[1]
        \renewcommand{\algorithmicrequire}{\textbf{Input:}}
        \renewcommand{\algorithmicensure}{\textbf{Output:}}
        \REQUIRE current state $x_0$, current control sequence (previous solution) $\{u^*\}^{(j)}$
        \ENSURE  re-planned trajectory $\{x^*\}^{(j+1)}$, re-planned control sequence $\{u^*\}^{(j+1)}$
        \\ \# \textit{Initialization}
         \STATE $\{x^\mathrm{r}\}=$ updateReferenceTrajectory()
        \STATE $\{u^*\}^{(j)}=$ stepControlSequenceForward($\{u^*\}^{(j)}$)
        \\ \# \textit{Loop process}
        \FOR {$i = 0$ to $\text{qp\_iterations}$}
        \STATE $l$ = generateTrajectoryLibrary($x_0$)
        \STATE $[\{x^\mathrm{c}\}, \{u^\mathrm{c}\}]=$ chooseCandidateFromLibrary($l$)
        \STATE $[\{\delta x^*\}, \{\delta u^*\}]=$ solveQP($\{x^\mathrm{c}\}, \{u^\mathrm{c}\}, \{x^\mathrm{r}\}$)
        \STATE $[\gamma, solved] =$ lineSearch($\{x^\mathrm{c}\}, \{\delta x^*\}, \{u^\mathrm{c}\}, \{\delta u^*\}$)
        \STATE ${u^c_k}=u^\mathrm{c}_k + \gamma \delta u^*_k, ~\forall k=0:T$
        \STATE $\{x^c\}=$ rollOutTrajectory($x_0, \{u^c\}$) 
        \ENDFOR
        \IF {solved}
        \STATE $\{x^*\}^{(j+1)}, \{u^*\}^{(j+1)}=\{x^c\},\{u^c\}$
        \ELSE
        \STATE $\{x^*\}^{(j+1)}, \{u^*\}^{(j+1)}=$ getStoppingTrajectory()
        \ENDIF
        \RETURN $\{x^*\}^{(j+1)}, \{u^*\}^{(j+1)}$
    \end{algorithmic} 
    \label{alg:full_algorithm}
\end{algorithm}






\textit{Trajectory library:} Our kinodynamic planner begins with selecting the best candidate trajectory from a trajectory library, which stores multiple initial control and state sequences.  The selected trajectory is used as initial solution for solving a full optimization problem.  The trajectory library can include:  1) the trajectory accepted in the previous planning iteration, 2) a stopping (braking) trajectory, 3) a geometric plan following trajectory, 4) heuristically defined trajectories (including v-turns, u-turns, and varying curvatures), and 5) randomly perturbed control input sequences.

\textit{QP Optimization:} Next, we construct a non-linear optimization problem with costs and constraints (\ref{eq:mpc_cost}--\ref{eq:mpc_state_constraints}).  We linearize the problem about the initial solution and solve iteratively in a sequential quadratic programming (SQP) fashion~\cite{nocedal2006numerical}.  Let $\{\hat{x}_k, \hat{u}_k\}_{k=0:T}$ denote an initial solution.  Let $\{\delta x_k, \delta u_k\}_{k=0:T}$ denote deviation from the initial solution.  We introduce the solution vector variable $X$:
\begin{equation}
    X = 
    \begin{bmatrix}
        \delta x_0^\Tr & \cdots & \delta x_T^\Tr & \delta u_0^\Tr & \cdots & \delta u_T^\Tr\\
    \end{bmatrix}^\Tr
\end{equation}
We can then write (\ref{eq:mpc_linear_start}--\ref{eq:mpc_linear_end}) in the form:
\begin{align}
    \text{minimize}\quad & \frac{1}{2}X^\Tr P X + q^\Tr X
    \label{eq:osqp_cost}\\
    \text{subject to}\quad & l\le AX \le u
    \label{eq:osqp_constraints}
\end{align}
where $P$ is a positive semi-definite weight matrix, $q$ is a vector to define the first order term in the objective function, $A$ defines inequality constraints and $l$ and $u$ provide their lower and upper limit. (See Appendix \ref{appendix:qp}.)
The next subsection (Section \ref{section:mpc_cost_constraints}) we describe these costs and constraints in detail.  This is a quadratic program, which can be solved using commonly available QP solvers such as our implementation uses the OSQP solver, which is a robust and highly efficient general-purpose solver for convex QPs \cite{osqp}.

\textit{Linesearch:} The solution to the SQP problem returns an optimized variation of the control sequence $\{\delta u_k^*\}_{k=0:T}$.  We then use a linesearch procedure to determine the amount of deviation $\gamma>0$ to add to the current candidate control policy $\pi$: $u_k=u_k + \gamma \delta u^*_k$. (See Appendix \ref{appendix:linesearch}.)

\textit{Stopping Sequence:} If no good solution is found from the linesearch, we pick the lowest cost trajectory from the trajectory library with no collisions.  If all trajectories are in collision, we generate an emergency stopping sequence to slow the robot as much as possible (a collision may occur, but hopefully with minimal energy).

\textit{Tracking Controller:} Having found a feasible and CVaR-minimizing trajectory, we send it to a tracking controller to generate closed-loop tracking behavior at a high rate (>100Hz), which is specific to the robot type (e.g. a simple cascaded PID, or legged locomotive controller).

\subsubsection{Optimization Costs and Constraints}\label{section:mpc_cost_constraints}
\textit{Costs:} Note that the cost \eqref{eq:mpc_cost} includes the CVaR risk.  To linearize the cost for representation in a QP, we compute the Jacobian and Hessian of $\rho$ with respect to the state $x$.  We efficiently approximate these terms via numerical differentiation.

\textit{Kinodynamic constraints:} Similar to the cost, we linearize the system dynamics \eqref{eq:mpc_dynamics} with respect to $x$ and $u$.  Depending on the dynamics model, this may be done analytically.

\textit{Control limits:}  The constraint function $g(u)$ in \eqref{eq:mpc_control_constraint} limits the range of the control inputs.  For example in the 6-state dynamics case, we limit maximum accelerations: $|a_x| < a^{\max}_x$, $|a_y| < a^{\max}_y$, and $|a_{\theta}| < a^{\max}_{\theta}$.

\textit{State limits:}  Within the constraint function $h(m, x)$ in \eqref{eq:mpc_state_constraints}, we encode velocity constraints: $|v_x| < v_{x}^{\max}$, $|v_y| < v_{y}^{\max}$, and $ |v_\theta| < v_{\theta}^{\max}$.  We also constrain the velocity of the vehicle to be less than some scalar multiple of the risk in that region, along with maximum allowable velocities:
\begin{align}
        |v_\theta| &< \gamma_\theta\,\rho(R_{k}) \\
        \sqrt{v_x^2 + v_y^2} &<  \gamma_v\,\rho(R_{k})
\end{align}
This constraint reduces the robot's kinetic energy in riskier situations, preventing more serious damage.

\begin{figure}[t!]
    \centering
    \includegraphics[width=150mm]{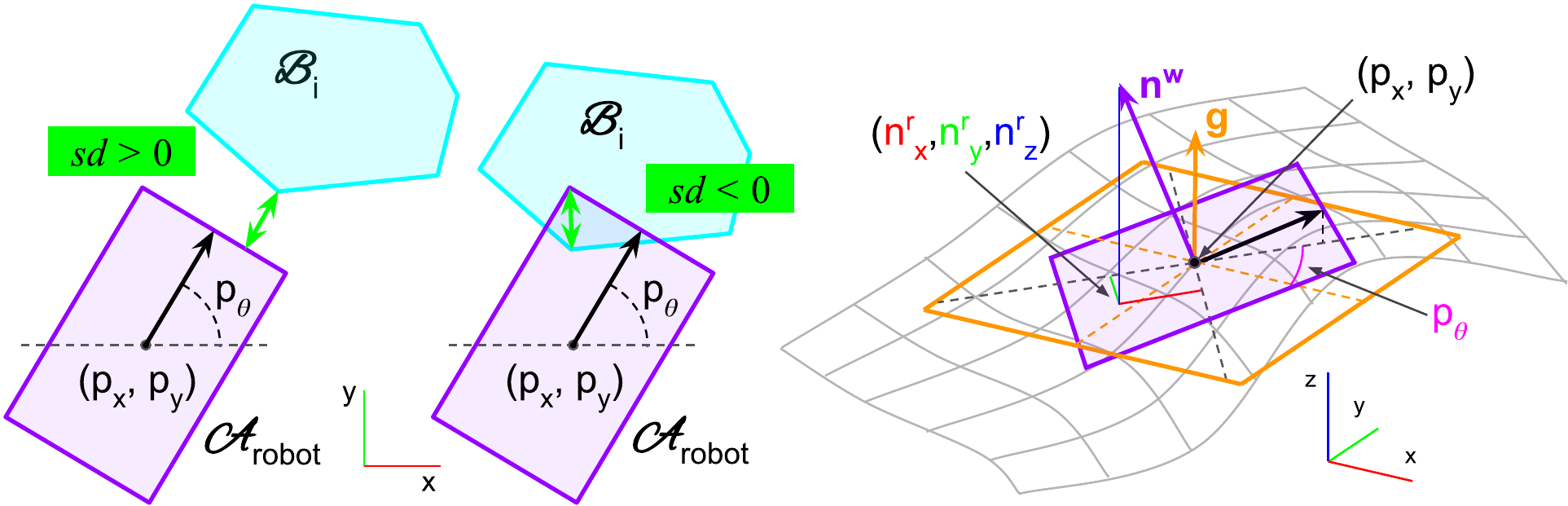}
    \caption{Left: Computing convex to convex signed distance function between the robot footprint and an obstacle.  Signed distance is positive with no intersection and negative with intersection.  Right: Robot pitch and roll are computed from the surface normal rotated by the yaw of the robot.  Purple rectangle is the robot footprint with surface normal $n^w$.  $\bm{g}$ denotes gravity vector, $n^r_{x,y,z}$ are the robot-centric surface normal components used for computing pitch and roll.}
    \label{fig:slope_angle}
\end{figure}

\textit{Position risk constraints:}  The function $h(m, x_k)$ also encodes constraints on robot position and orientation to prevent the robot from hitting obstacles.  The general form of this constraint is:
\begin{align}\label{eq:mpc_position_const}
    \rho(R_k) < \rho^{\max}.
\end{align}
To formulate the constraint~\eqref{eq:mpc_position_const}, we locate areas on the map where the risk $\rho$ is greater than the maximum allowable risk. These areas are marked as obstacles, and are often highly non-convex. To obtain a convex and tractable approximation of this highly non-convex constraint, we decompose obstacles into non-overlapping 2D convex polygons, and create a signed distance function which determines the minimum distance between the robot's footprint (also assumed to be a convex polygon) and each obstacle \cite{schulman2014motion}.  Let $\mathcal{A}, \mathcal{B}\subset \mathbb{R}^2$ be two convex sets, and define the distance between them as:
\begin{align}
    \mathrm{dist}(\mathcal{A}, \mathcal{B}) = \inf\{\|T\| ~ | ~ (T+\mathcal{A})\cap\mathcal{B} \neq \emptyset\}
\end{align}
where $T$ is a translation.  When the two sets are overlapping, define the penetration distance as:
\begin{align}
    \mathrm{penetration}(\mathcal{A}, \mathcal{B}) = \inf\{\|T\| ~ | ~ (T+\mathcal{A})\cap\mathcal{B} = \emptyset\}
\end{align}
Then we can define the signed distance between the two sets as:
\begin{align}
    \mathrm{sd}(\mathcal{A}, \mathcal{B})=\mathrm{dist}(\mathcal{A},\mathcal{B}) - \mathrm{penetration}(\mathcal{A},\mathcal{B})
\end{align}
We then include within the function $h(m,x_k)$ a constraint to enforce the following inequality:
\begin{align}
    \mathrm{sd}(\mathcal{A}_{\mathrm{robot}},\mathcal{B}_i) > 0 \quad \forall i \in \{0,\cdots,N_{\mathrm{obstacles}}\}.
\end{align}
Note that the robot footprint $\mathcal{A}_{\mathrm{robot}}$ depends on the current robot position and orientation: $\mathcal{A}_{\mathrm{robot}}(p_x, p_y, p_{\theta})$, while each obstacle $\mathcal{B}_i(m)$ is dependent on map information (See Figure \ref{fig:slope_angle}).

\textit{Orientation constraints:}
To constrain the robot's orientation on sloped terrain so as to prevent the robot from rolling over or performing dangerous maneuvers, we add constraints to the function 
$h(m, x_k)$ which limit the roll and pitch of the robot as it settles on the surface of the ground.  Denote the robot's position as $p=[p_x,p_y]^\intercal$ and the position/yaw as $s=[p_x,p_y,p_\theta]^{\intercal}$.  Let $\psi$ and $\phi$ denote the robot's pitch and roll in its body frame.  Let $\omega=[\psi, \phi]^\intercal$.  The constraint will have the form $|\omega| \prec \omega^{\max}$. Let the surface normal vector at point $p$ be denoted by $n^w=[n^w_x,n^w_y,n^w_z]^\intercal$, in the world frame.  Let $n^r = [n^r_x,n^r_y,n^r_z]^\intercal$, be the surface normal in the body frame, where we rotate by the robot's yaw: $n^r = R_\theta n^w$ (see Figure \ref{fig:slope_angle}), where rotation matrix $R_\theta$ represents a rotation by the angle $\theta$ about the world $z$ axis.
Then, we define the robot pitch and roll as $\omega = g(n^r)$ where:
\begin{align}
\omega = g(n^r) =
\begin{bmatrix}
    \mathrm{atan2}(n^r_x,n^r_z)\\
    -\mathrm{atan2}(n^r_y,n^r_z)
\end{bmatrix}
\end{align}
Creating a linearly-constrained optimization problem requires a linear approximation of the constraint: 
\begin{equation}
    \label{eq:orient_con}
    |\nabla_s\omega(s)\delta s + \omega(s)| < \omega^{\max}
\end{equation}
The linearization is realized by finding the gradients of the orientation constraints with respect to position and yaw separately (See Appendix \ref{appendix:grad}).

\textit{Box Constraint:} Note that if $\delta x$ and $\delta u$ are too large, linearization errors will dominate.  To mitigate this effect we also include box constraints within \eqref{eq:mpc_control_constraint} and \eqref{eq:mpc_state_constraints} to maintain a bounded deviation from the initial solution: $|\delta x| < \epsilon_x$ and $|\delta u| < \epsilon_u$.

\textit{Adding Slack Variables:}  To further improve the feasibility of the optimization problem we introduce auxilliary slack variables for constraints on state limits, position risk, and orientation.  For a given constraint $h(x)>0$ we introduce the slack variable $\epsilon$, and modify the constraint to be $h(x)>\epsilon$ and $\epsilon<0$.  We then penalize large slack variables with a quadratic cost: $\lambda_\epsilon \epsilon^2$.  These slack variables are incorporated into the QP problem \eqref{eq:osqp_cost} and \eqref{eq:osqp_constraints}.


\subsection{Dynamic Risk Adjustment}

The CVaR metric allows us to dynamically adjust the level and severity of risk we are willing to accept. Selecting low $\alpha$ reverts towards using the mean cost as a metric, leading to optimistic decision making while ignoring low-probability but high cost events. Conversely, selecting a high $\alpha$ leans towards conservatism, reducing the likelihood of fatal events while reducing the set of possible paths. We adjust $\alpha$ according to two criteria:  1)  Mission-level states, where depending on the robot's role, or the balance of environment and robot capabilities, the risk posture for individual robots may differ.  2)  Recovery Behaviors, where if the robot is trapped in an unfavorable condition, by gradually decreasing $\alpha$, an escape plan can be found with minimal risk.  These heuristics are especially useful in the case of risk-aware planning, because the feasibility of online nonlinear MPC is difficult to guarantee.  When no feasible solution is found for a given risk level $\alpha$, a riskier but feasible solution can be quickly found and executed.
\subsection{Risk-Driven Gait Selection}\label{sec:gait_switch}
We considered two platforms for testing our traversability framework - a Clearpath Husky (wheeled robot) and Boston Dynamics Spot (quadrupedal legged robot). In the case of the Spot, multiple locomotion gaits are available, namely, \textit{walking gait} (standard operation, most stable), \textit{stair climbing gait} (reduces the robot speed and pitches the robot to better see the stairs while descending), \textit{crawling gait} (three feet touch the ground at all times, the most stable gait)\footnote{Further descriptions of the gaits can be found here: \url{https://support.bostondynamics.com/s/article/Operating-Spot}}. Furthermore, the height of the robot can also be dynamically adjusted; the height reduction to fit in small spaces is henceforth called the \textit{crouch gait}. These locomotion gaits provide mobility for the Spot in areas that would not be traversable with the standard walking gait. 

\begin{itemize}
    \item \ph{Stair gait selection} 
    Stair detection was performed using geometric analysis of LiDAR and camera disparity pointcloud data.  To detect stairs, we use a plane fitting method similar to \cite{schwarze2015stair}.  We compute normals and perform clustering to identify contiguous surfaces in the pointcloud of similar slope.  We then isolate flat and vertical planes to detect potential stair surfaces.  After finding candidate stair surfaces, the larger stair direction and overall plane is determined.  Finally, the intersection of the ground plane that the robot currently sits on and the plane of the stairs is used to determine the start of the stairs. Once the start of the stairs (either up or down) is determined, the robot can align itself to this location and begin a stair walking gait.
    
\begin{figure*}[ht]
    \centering
    \includegraphics[width=0.99\textwidth]{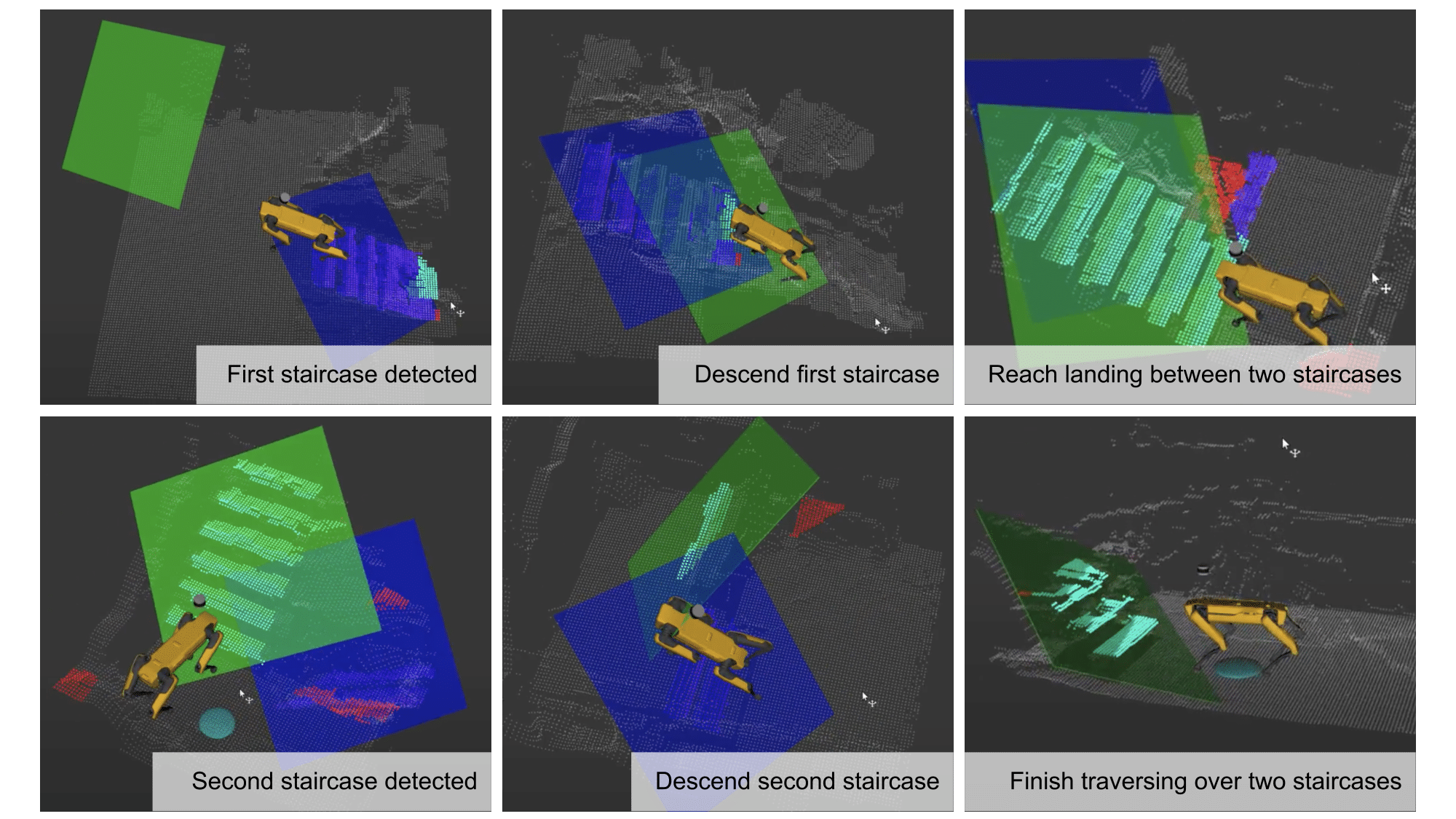}
    \caption{Stair detection with plane-fitting method}
    \label{fig:stair_behavior}
\end{figure*}
   
    \item \ph{Crouch gait through slip detection}
A learning-based slip prediction module was implemented in~\cite{dey2022prepare} that leveraged the joint state information of the quadruped at each time instant. We train an ensemble model to predict the probability of slip, $p(slip)$, at each time instant, based on the robot limb kinematics and kinetics. Slip-annotations from multiple field trials across different terrains (e.g., rocky, sandy, muddy terrains) formed the ground truth for the model training. During slip prediction using the trained model, if the $p(slip) > 0.5$, the \textit {crawling gait} mode is enabled for the robot to traverse the potentially slippery terrain. For more details on the implementation of the slip-predictive model, see~\cite{dey2022prepare}.

\begin{figure*}[ht]
    \centering
    \includegraphics[width=0.65\textwidth]{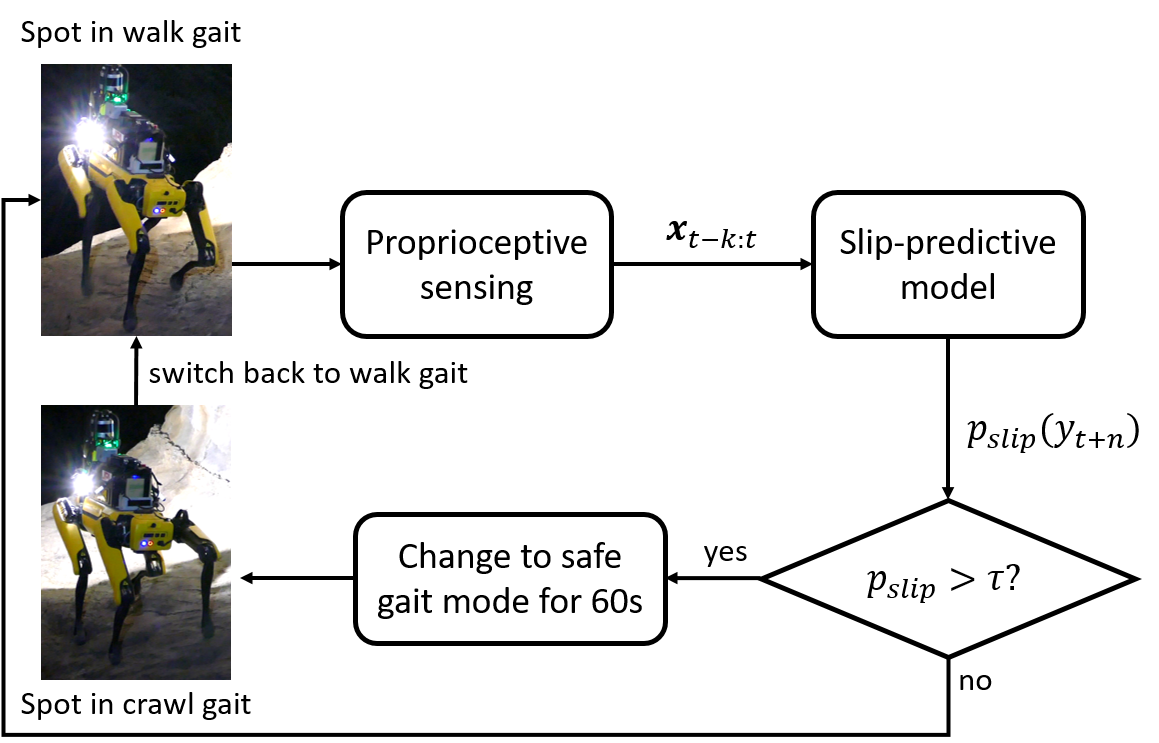}
    \caption{Gait change through slip detection}
    \label{fig:slip_detection}
\end{figure*}


    \item \ph{Crouch behavior in low ceiling areas} If the robot detects a region with low ceiling that is within 16 cm of the robot height (the limit to which the robot can bend), the robot enables \textit{crouch gait} that reduces its height of the robot by 16 cm. Once the robot passes through the low ceiling area, it switches back to the \textit{walking gait}.
    
    \begin{figure*}[ht]
    \centering
    \includegraphics[width=135mm]{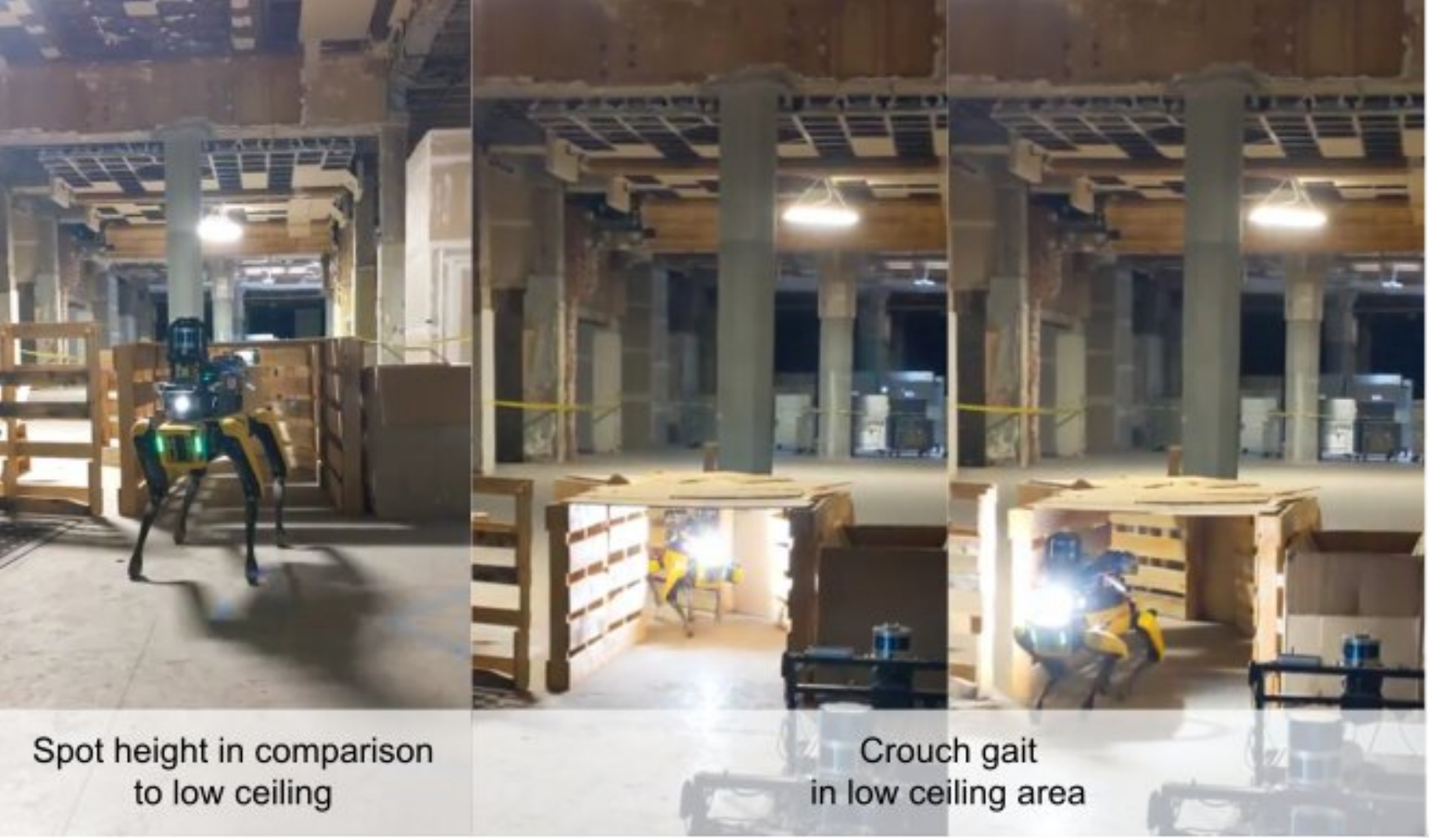}
    \caption{Crouch behavior in low ceiling areas}
    \label{fig:crouch_behavior_LA}
\end{figure*}
   
\end{itemize}
\subsection{Recovery Behaviors}
Recovery behaviors ensure fast, reactive actions to hazardous, mission-ending situations such that the robot can recover quickly and safely. In the case that a recovery behavior is activated, the robot temporarily pauses following the commands from the geometric and kinodynamic planning pipeline to get to a safe state from where it can resume the hierarchical planning pipeline (see~\autoref{fig:system}). 

\ph{Tilt recovery}
The tilt recovery behavior tracks the odometry received from the state estimation pipeline for the wheeled robot. The odometry is a combination of wheel odometry and inertial measurements. It consists of the pose and twist measurements. If the robot's pitch exceeds a threshold, the robot backtracks at a low speed along the same path that got the robot in the lethal pitch state.

\begin{figure*}[ht]
    \centering
    \includegraphics[width=0.99\textwidth]{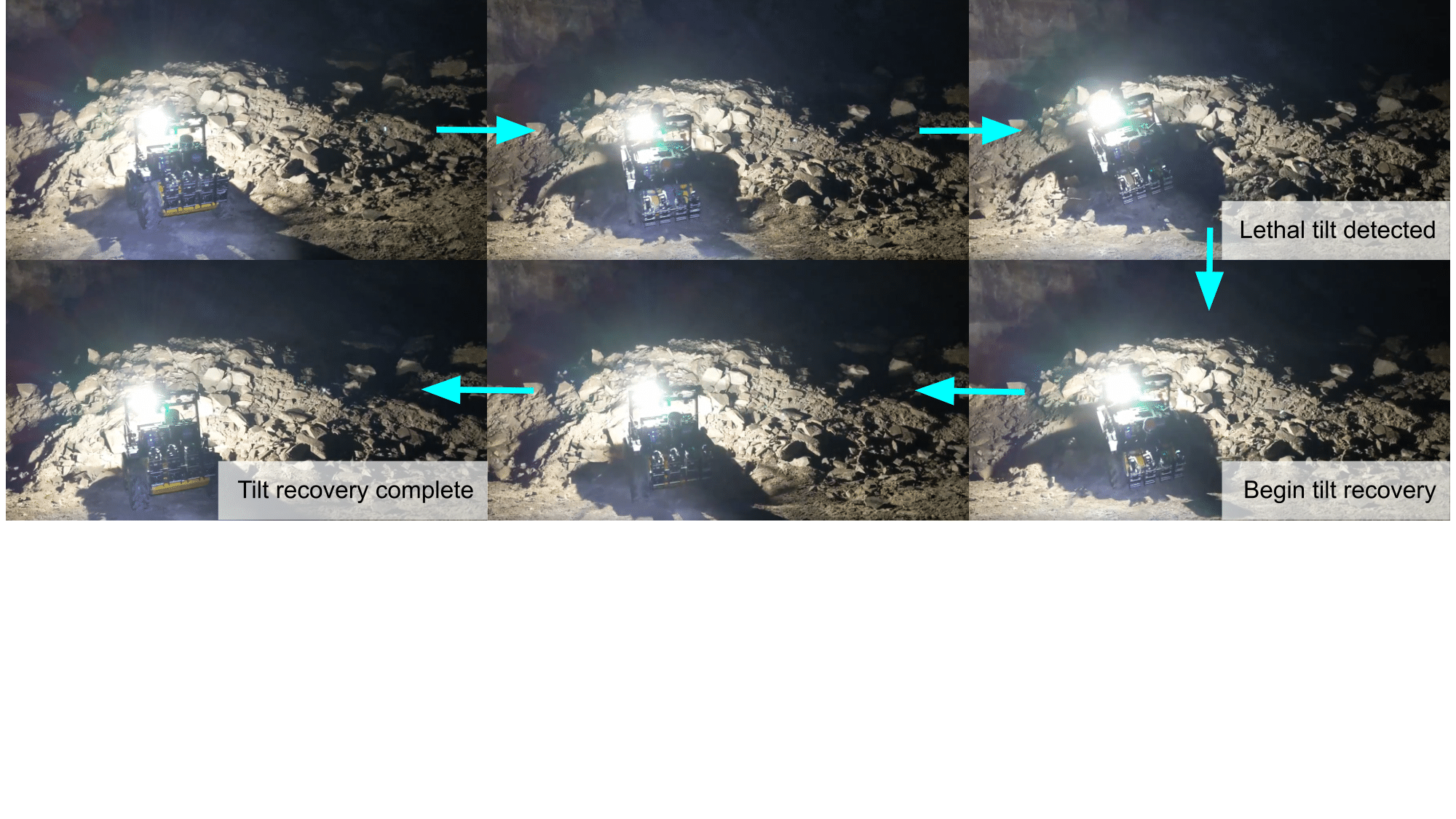}
    \vspace{-35mm}
    \caption{Tilt recovery behavior when the robot pitch exceeds the safe limit.}
    \label{fig:tilt_recovery_KU}
\end{figure*}

\ph{Wiggle behavior}
We monitor the robot's velocity obtained from wheel and inertial odometry to ensure that the robot is not stuck during mission. If the robot does not move for a certain time period while the MPC provides robot movement commands, the robot is considered to be stuck. In this case, we deploy a wiggle behavior that moves the wheels of the robot in such a way that, unintruded, the robot would travel in a figure-8 pattern. The allows the robot to wiggle back and forth till it is not stuck anymore. If the wiggle behavior fails to ensure recovery, the robot is considered stuck and the mission ends.

\ph{Escape lethal}
If the robot finds itself trapped in a lethal zone for some period of time, it is useful to have a behavior which attempts to escape the high risk regions with minimum risk.  This is accomplished in the following way.  First, a goal location is chosen which is not in a lethal area, close to the robot, and which minimizes the amount of lethal area the robot would need to traverse to reach the goal in a straight line.  Next, the risk threshold of the traversability cost maps are changed, to allow traversal through high risk map cells which would normally be treated as constraints or obstacles.  Then, the goal is sent to the global and local planners, and executed on the robot.  If the goal is reached, the recovery behavior is ended, but if the goal is still not reachable, the risk threshold can be further adjusted to allow more and more risky behavior in an attempt to escape.  In this way, the CVaR risk threshold is useful for adapting the risk profile of the robot in real-time.


\section{Experiments}

In this section, we report the experimental and field performance of STEP. We first present a comparative study between different adjustable risk thresholds in simulation on a wheeled differential drive platform.  Then, we demonstrate real-world performance using a wheeled robot deployed in an abandoned subway filled with clutter, and a legged platform deployed in a lava tube environment.

\subsection{Simulation Study}

\begin{figure}
    \centering
    \includegraphics[width=0.85\textwidth]{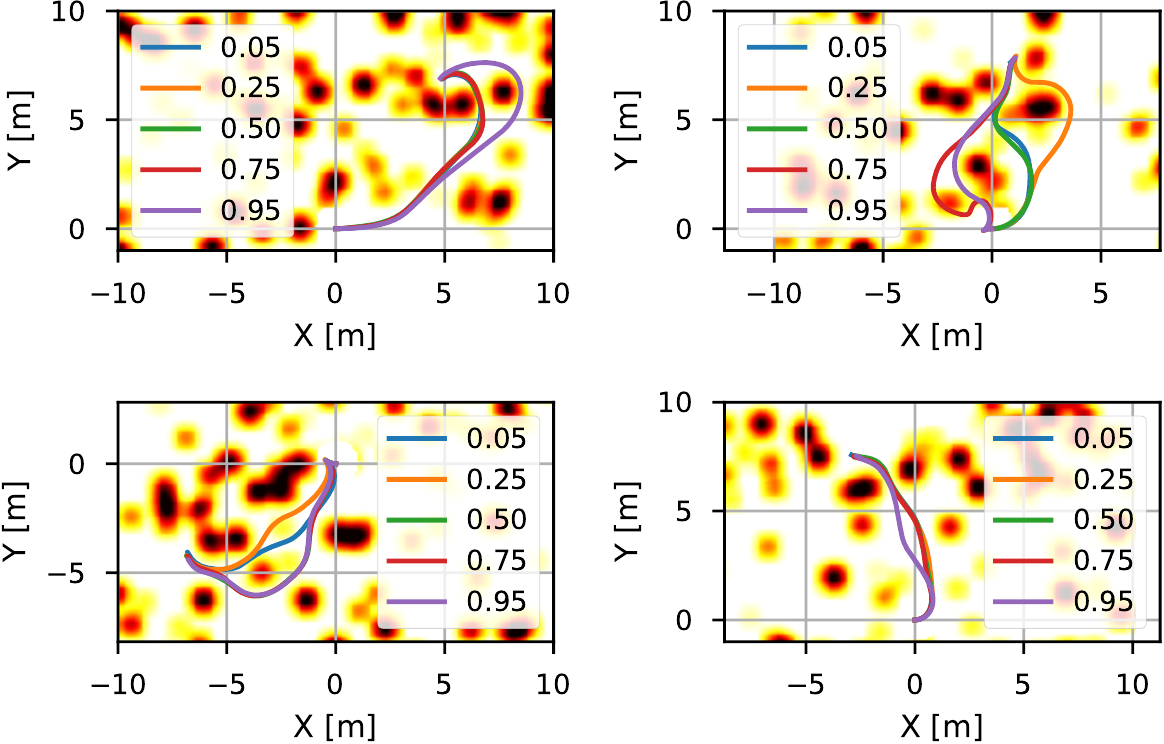}
    \caption{Path distributions from four simulated runs. The risk level $\alpha$ spans from 0.1 (close to mean-value) to 0.95 (conservative). Smaller $\alpha$ typically results in a shorter path, while larger $\alpha$ chooses statistically safe paths.}
    \label{fig:sim_pathdist}
\end{figure}

\begin{figure}
    \centering
    \includegraphics[width=0.75\textwidth]{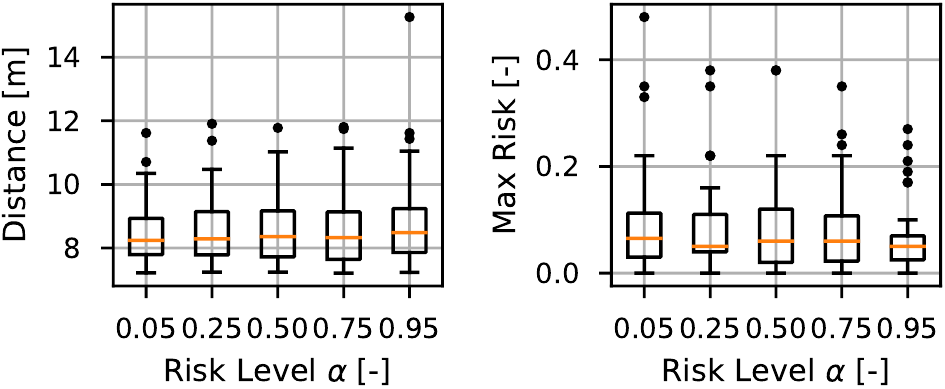}
    \caption{Distance vs risk trade-off from 50 Monte-Carlo simulations. Left: Distributions of path distance. Right: Distributions of max risk along the traversed paths.  Box plot uses standard quartile format and dots are outliers.}
    \label{fig:sim_tradeoff}
\end{figure}

To assess statistical performance, we perform 50 Monte-Carlo simulations with randomly generated maps and goals. These results were also included in our previous work~\cite{fan2021step}. Random traversability costs are assigned to each grid cell. The following assumptions are made: 1) no localization error, 2) no tracking error, and 3) a simplified perception model with artificial noise. We give a random goal 8\,m away and evaluate the path cost and distance.  We use a differential-drive dynamics model (no lateral velocity).

We compare STEP using different $\alpha$ levels.  Figure \ref{fig:sim_pathdist} shows the distribution of paths for different planning configurations. The optimistic (close to mean-value) planner $\alpha=0.05$ 
typically generates shorter paths, while the conservative setting $\alpha=0.95$ makes long detours to select statistically safer paths. The other $\alpha$ settings show distributions between these two extremes, with larger $\alpha$ generating similar paths to the conservative planner and smaller $\alpha$ generating more time-optimal paths. Statistics are shown in Figure \ref{fig:sim_tradeoff}.

\subsection{Hardware Results}
We deployed STEP on two different robots (wheeled and legged) in different challenging environments (an abandoned subway, a lava tube, a limestone mine, and a combination of different subterranean environments in the DARPA Subterranean Challenge). The robot was equipped with custom sensing and computing units, and driven by JPL's NeBula autonomy software \cite{AliNeBula21}.  3 Velodyne VLP-16s were used for collecting LiDAR data.  Localization was provided onboard by a LiDAR-based SLAM solution \cite{Palieri2020,Ebadi2020}.  The entire autonomy stack runs on an Intel Core i7 CPU. The typical CPU usage for the traversability stack is about a single core.  

\subsubsection{Previous Results}
\begin{figure*}[ht]
    \centering
    \includegraphics[width=0.99\textwidth]{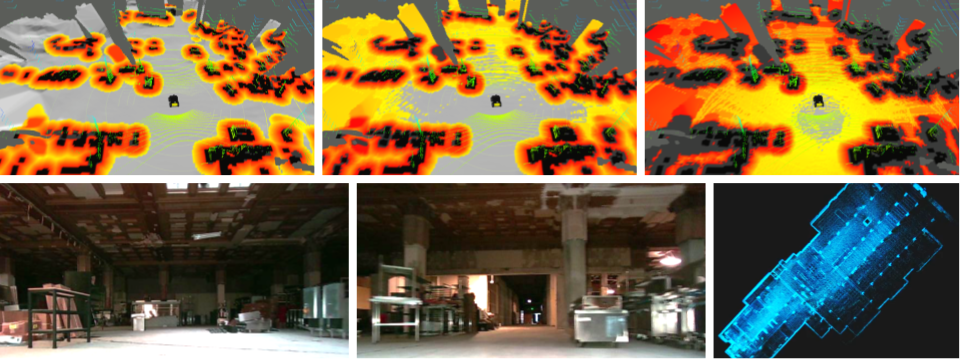}
    \caption{Traversability analysis results for Husky in an abandoned subway experiment. Top left to right: Risk maps at three varying risk levels: $\alpha = 0.1, ~0.5, ~0.9$, respectively. Colors correspond to CVaR value (white: safe ($r<=0.05$), yellow to red: moderate ($0.05 < r <= 0.5$), black: risky ($r > 0.5$)).  Also shown are the most recent LiDAR measurements (green points).  Bottom left and middle:  Front and right on-board cameras observing the same location.  Bottom right:  Completed top-down map of the environment after autonomous exploration.  Bright dots are pillars, which are visible in the camera images. The effect of these risk analyses results in intuitive outcomes - for example, a low pile of metal, while probably traversable, should be avoided if possible.  When the region has inadequate sensor coverage, the risk will be high.  When the robot is closer and the sensor coverage is good, then the CVaR cost will decrease, yielding a more accurate risk assessment.  This results in more efficient and safer planning when compared to deterministic methods.}
    \label{fig:husky_subway}
\end{figure*}

\begin{figure*}[ht]
    \centering
    \includegraphics[width=145mm]{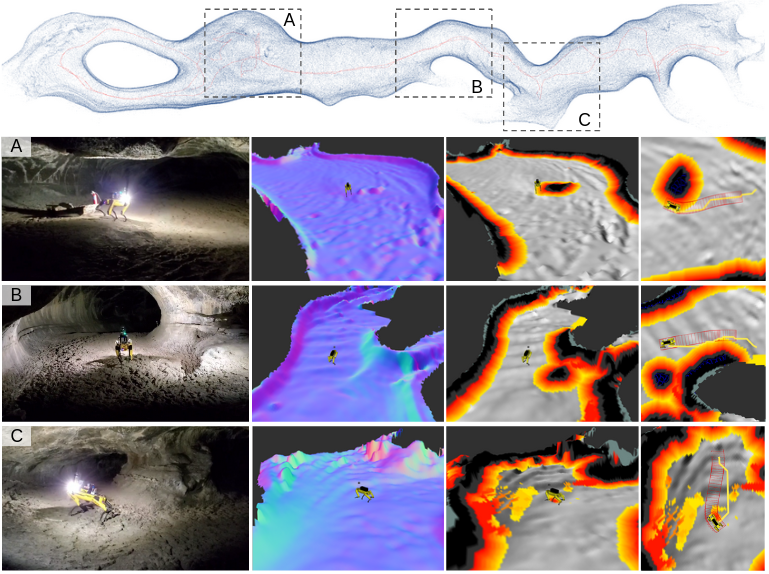}
    \caption{Traversability analysis results for the Valentine Cave experiment. From left to right: Third-person view, elevation map (colored by normal direction), risk map (colored by risk level. white: safe ($r<=0.05$), yellow to red: moderate ($0.05 < r <= 0.5$), black: risky ($r > 0.5$)), and planned geometric/kinodynamic paths (yellow lines/red boxes). }
    \label{fig:valentine_result}
\end{figure*}

Earlier versions of STEP (without semantic and confidence-aware risk sources, gait adaptations, and the entire suite of recovery behaviors) were tested in multiple field locations. For the sake of completeness, we have included these results, published in~\cite{fan2021step}, from:
\begin{enumerate}
    \item Abandoned subway filled with industrial clutter in Downtown Los Angeles, CA in Figure~\ref{fig:husky_subway},
    \item Valentine Cave in Lava Beds National Monument, Tulelake, CA in Figure~\ref{fig:valentine_result}
\end{enumerate}

\subsubsection{Results from the Kentucky Underground and LA Subway}
We tested the various new parts of our traversability pipeline during fully autonomous runs in our tests at Kentucky Underground Storage (KU), Wilmore, KY. The results of the cost map for confidence-based and semantics-based risk are seen in Figure~\ref{fig:neg_obs_risk} and Figure~\ref{fig:water_risk}. We also tested the tilt recovery behavior on a pile of rubble in KU by forcing the robot into a hazardous tilt position, the robot was able to recover from this risky configuration as seen in Figure~\ref{fig:tilt_recovery_KU}.

We also tested new results in an abandoned subway with multiple levels filled with industrial clutter in Downtown Los Angeles, CA. For example, the LA Subway envirnoment served as a test bed for the stair gait and crouch gait behaviors that are shown in Figure~\ref{fig:stair_behavior} and Figure~\ref{fig:crouch_behavior_LA} respectively.

\subsubsection{Results from the DARPA Subterranean Challenge}
\begin{figure*}[ht]
    \centering
    \includegraphics[width=0.99\textwidth]{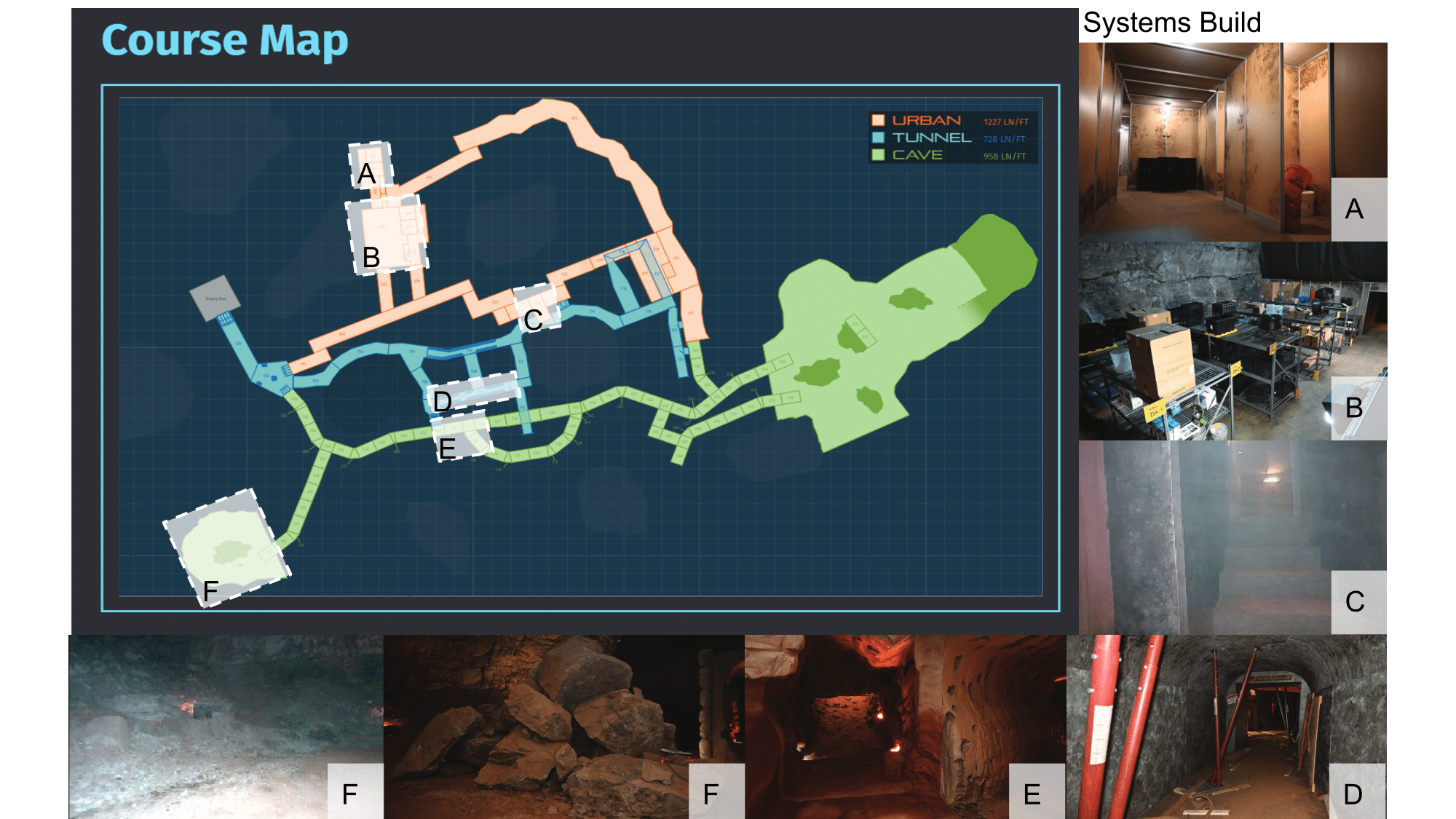}
    \caption{The layout of the competition course with snapshots of the systems build of the areas traversed by the robots that have interesting terrain and high difficulty.}
    \label{fig:finals_map}
\end{figure*}

\begin{figure}
    \centering
\begin{subfigure}[b]{0.99\textwidth}
    \centering
    \includegraphics[width=0.99\textwidth]{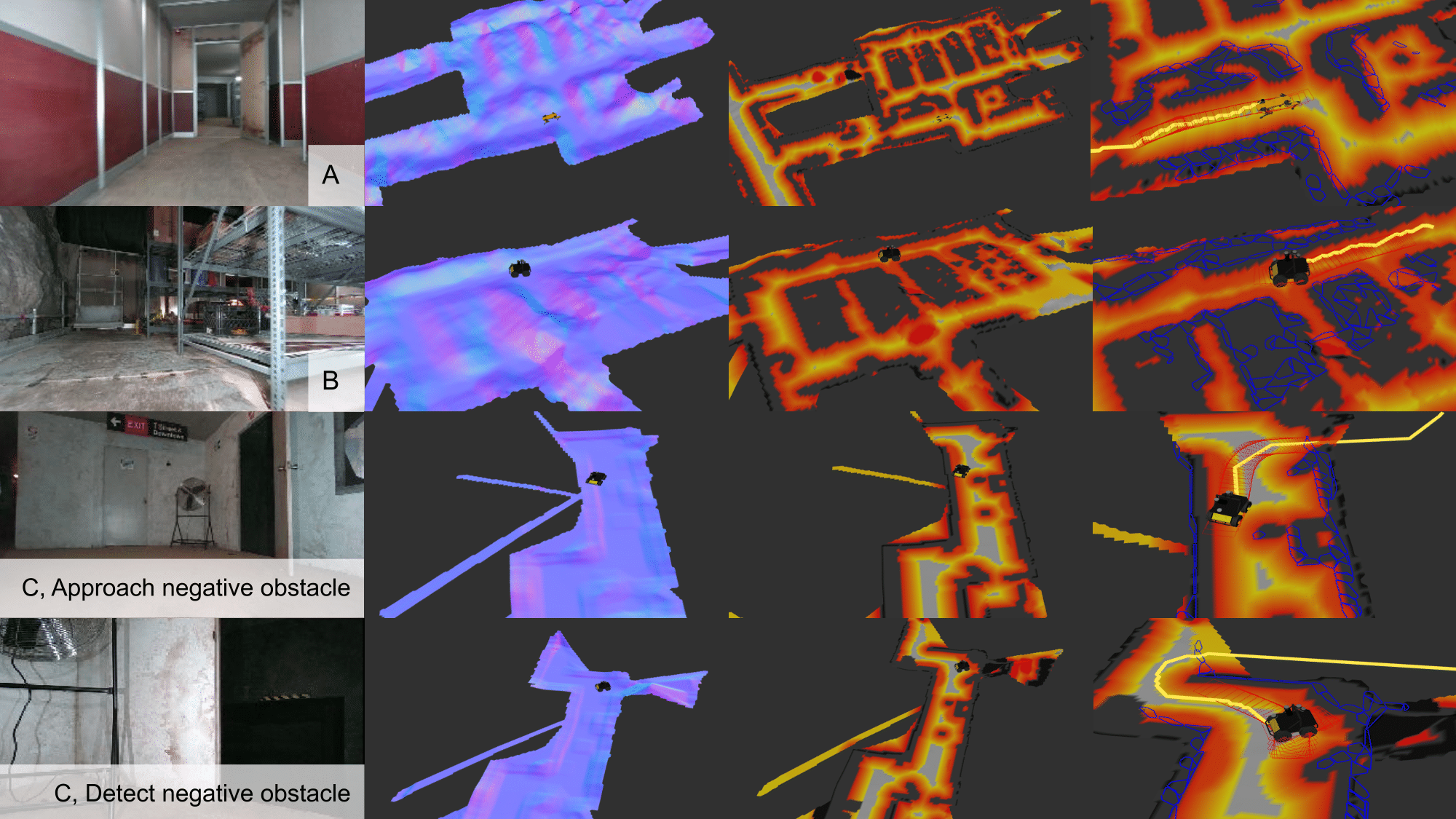}
\end{subfigure}\\
\begin{subfigure}[b]{0.99\textwidth}
    \centering
    \vspace{-5mm}
    \includegraphics[width=0.99\textwidth]{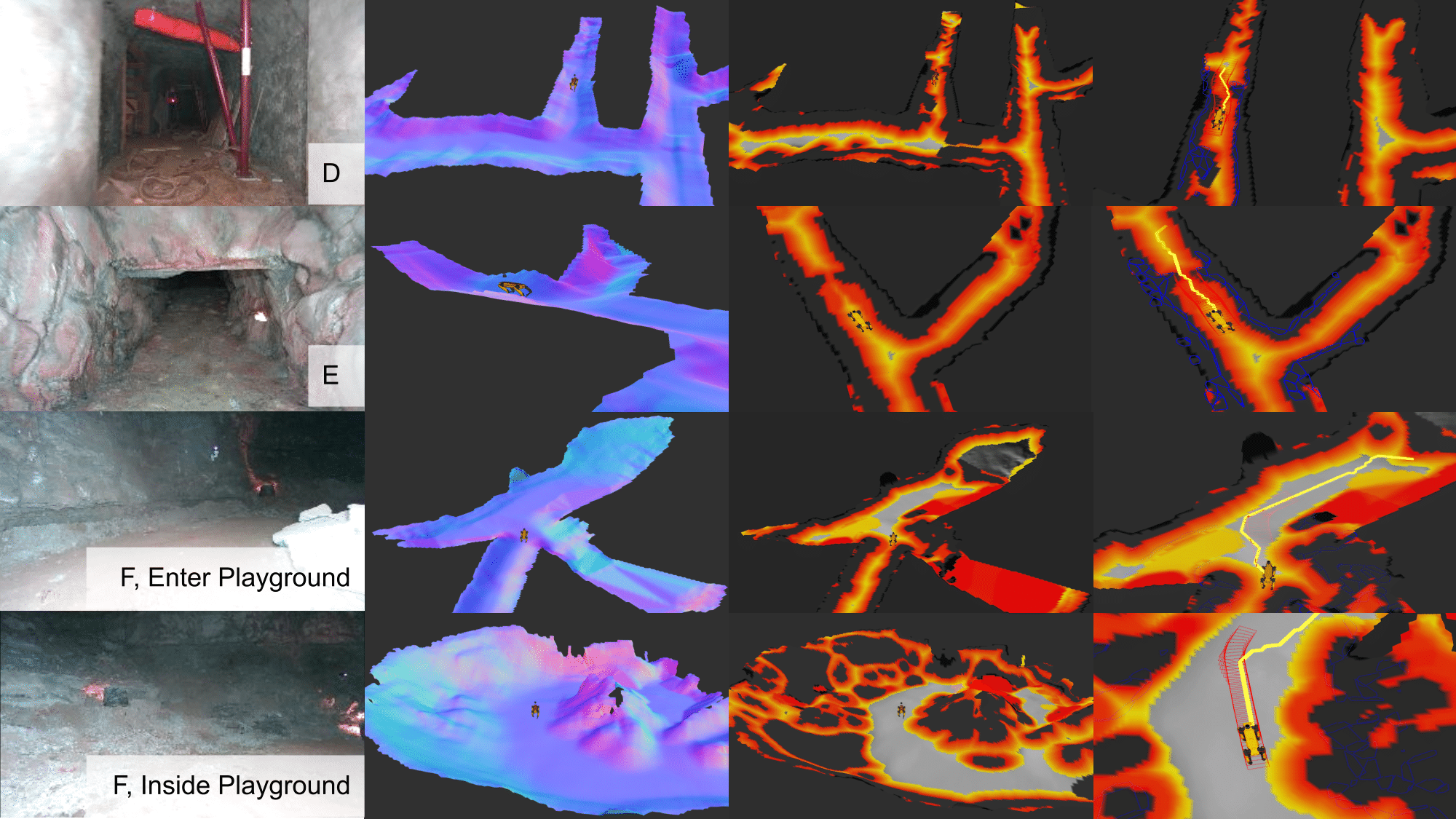}
\end{subfigure}
 \caption{Traversability analysis results for the DARPA Subterranean Challenge. From left to right: robot front camera view, elevation map (colored by normal direction), risk map (colored by risk level- white: safe ($r<=0.05$), yellow to red: moderate ($0.05 < r <= 0.5$), black: risky ($r > 0.5$)), and planned geometric/kinodynamic paths (yellow lines/red boxes).}
    \label{fig:finals_results}
\end{figure}

The traversability framework was deployed in the DARPA Subterranean Challenge held in Louisville Mega Cavern, Louisville, KY. The course consisted of 3 different environments - tunnel, urban, and cave. Figure~\ref{fig:finals_map} shows the course map and different sections that the robot explored. The sections showcased in Figure~\ref{fig:finals_map} have a high difficulty rating as listed in the course layout guide posted by DARPA\footnote{The course guide can be accessed here: \url{https://bitbucket.org/subtchallenge/finals_ground_truth/src/master/course_design/Finals_Course_Callouts.pdf}}. The terrain challenges of each of the sections are listed below:
\begin{description}
\item[Region A] An office-like area that consists of narrow corridors ($5'\times8'$) and small rooms ($8'\times8'$) for the robots to explore. 
\item[Region B] A warehouse-like area with a lot of shelving and clutter imitating an industrial warehouse after an earthquake. 
\item[Region C] A connection between the urban and tunnel part of the course. The stairs act as a negative obstacle for the wheeled robots. The presence of this negative obstacle in a narrow corridor makes the drop harder to detect.
\item[Region D] A constrained passage with ground, wall, and ceiling obstacles - with vertical pipes and debris.
\item[Region E] A narrow cave opening mimicking a region that humans have to crawl through. The ground slopes upwards and the ceiling slopes downwards creating issues for ground segmentation and low-ceiling detection.
\item[Region F] A small limestone cave with rubble and loose rock piles.
\end{description}
Figure~\ref{fig:finals_results} shows the results of the traversability analysis and the geometric and kinodynamic plans of the robot in the different scenarios listed above. We see that the robot is able to plan safe paths in all of these scenarios and successfully navigate the different terrain hazards.

\section{Discussion}
We presented our approach to Stochastic Traversability Evaluation and Planning and the performance of our method in the DARPA Subterranean Challenge. The method was successfully able to account for uncertainty in the apriori unknown environment and navigate various types of terrain. In this extension of~\cite{fan2021step}, we accounted for semantic and confidence-aware risk sources in addition to the previously presented geometric risk sources. Additionally, we allowed for changes in gaits of quadrupedal robots based on the terrain risk observed and information obtained from priprioceptive sensors. We added a large suite of recovery behaviors to robustify our traversability stack in case of unaccounted for but mission-ending situations that the rest of the framework cannot account for.

There are many avenues of improvement of our framework. STEP accounts for traversability risk arising from static environments, but it does not make predictions for dynamic obstacles. Hence, the robot cannot plan reactive motions to avoid falling debris or account for sudden movement of rubble under the robot. The traversability analysis is also the same for all speeds of the robot. Dynamic traversability analysis will allow for a larger range of motion. The LIDAR intensity-based semantic terrain analysis provides us with a coarse classification of muddy terrain vs regular terrain but using a vision-based semantic classification in addition to the LIDAR intensity information will provide a richer understanding of the surface over which the robot is traversing.

\clearpage
\newpage

\section*{Acknowledgement}
The work is performed at the Jet Propulsion Laboratory, California Institute of Technology, under a contract with the National Aeronautics and Space Administration (80NM0018D0004), and Defense Advanced Research Projects Agency (DARPA).

\printbibliography

\clearpage
\newpage
\appendix
\section{Dynamics model for differential drive}
\label{appendix:dyn}
For a simple system which produces forward velocity and steering, (e.g. differential drive systems), we may wish to specify the state and controls as:
\begin{align}
    x = [p_x, p_y, p_\theta, v_x]^{\intercal} \\
    u = [a_x, v_{\theta}]^{\intercal}
\end{align}
For example, the dynamics $x_{k+1} = f(x_k,u_k)$ for a simple differential-drive system can be written as:
\begin{align}
    x_{k+1} = x_k + \Delta t \begin{bmatrix}
        v_x \cos(p_\theta)\\
        v_x \sin(p_\theta)\\
        \gamma v_x + (1-\gamma) v_{\theta}\\
        a_x
    \end{bmatrix}
\end{align}
where $\gamma\in[0,1]$ is a constant which adjusts the amount of turning-in-place the vehicle is permitted.

\section{Dynamic risk metric using CVaR for Normal distributions}
\label{appendix:cvar}
\begin{equation}
\begin{aligned}
    J(x_0, \pi) &= R_0 + \rho_0\big( R_1 + \rho_{1}\big(R_2 + \dotsc + \rho_{T-1}(R_{T})\big)\big) \\
    &= R_0 + \rho\Bigg( R_1 + \rho\Big(R_2 + \dotsc + \rho(R_{T-1} + \\ &\qquad\mu_T + \sigma_T \frac{\varphi(\boldsymbol{\Phi}^{-1}(\alpha))}{1-\alpha})\Big)\Bigg) \\
    &= R_0 + \rho\Bigg( R_1 + \rho\Big(R_2 + \dotsc + \rho(R_{T-2} + \\  &\qquad\mu_{T-1} + \mu_T + (\sigma_{T-1} + \sigma_T) \frac{\varphi(\boldsymbol{\Phi}^{-1}(\alpha))}{1-\alpha})\Big)\Bigg) \\
    & \vdots \\
    &= R_0 + \sum_{i=1}^{T}\Bigg(\mu_i + \sigma_i \frac{\varphi(\boldsymbol{\Phi}^{-1}(\alpha))}{1-\alpha}\Bigg)\\
    &= \sum_{i=0}^{T}\rho(R_i)\nonumber
\end{aligned}
\end{equation}

\section{Kinodynamic Planning Diagram}
\label{appendix:traj_fig}
\begin{figure}[ht!]
    \centering
    \includegraphics[width=150mm]{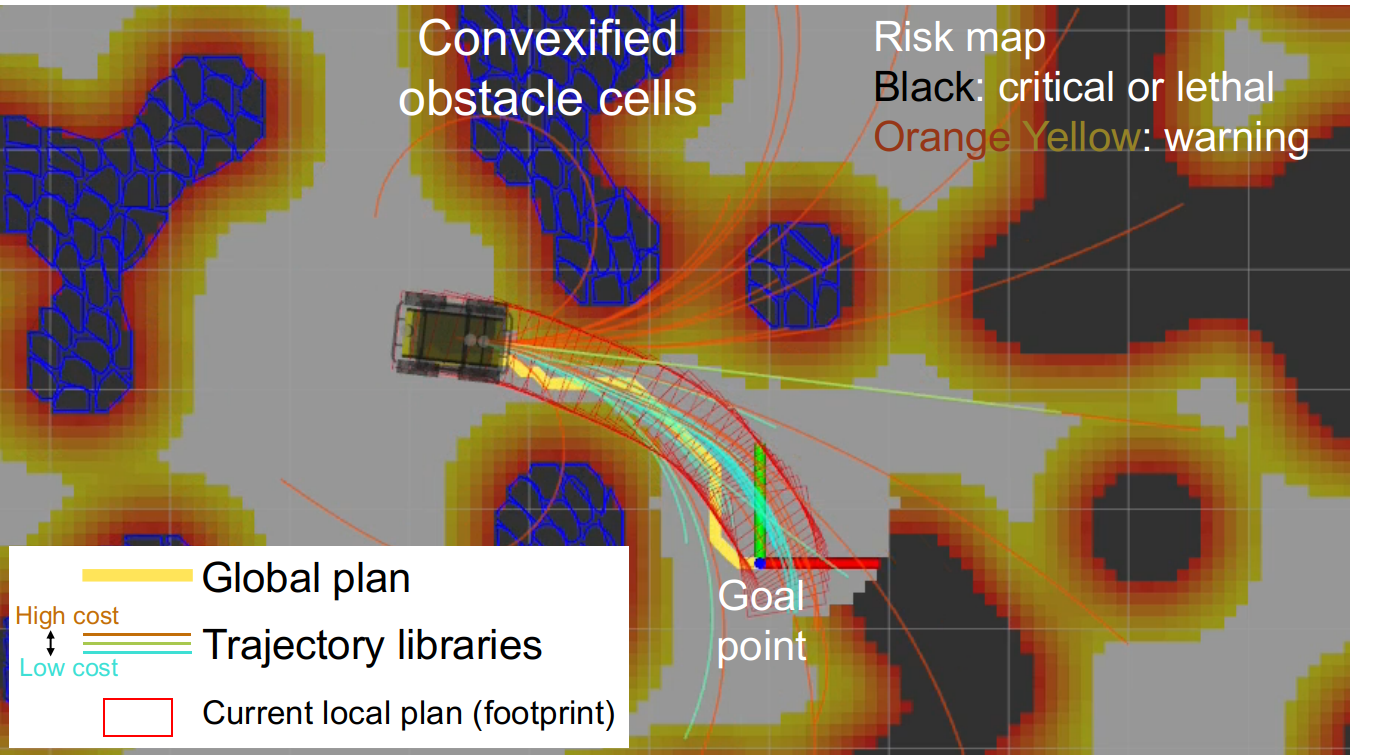}
    \vspace{5mm}
    \caption{Diagram of kinodynamic MPC planner, which begins with evaluating various paths within a trajectory library.  The lowest cost path is chosen as a candidate and optimized by the QP solver.}
    \label{fig:traj_lib}
\end{figure}

See~\autoref{fig:traj_lib}.

\section{Gradients for Orientation Constraint}
\label{appendix:grad}
We describe in further detail the derivation of the orientation constraints.  Denote the position as $p=[p_x,p_y]^\intercal$ and the position/yaw as $s=[p_x,p_y,p_\theta]$.  We wish to find the robot's pitch $\psi$ and roll $\phi$ in its body frame.  Let $\omega=[\psi, \phi]^\intercal$.  The constraint will have the form $|\omega(s)| <= \omega_{max}$.  At $p$, we compute the surface normal vector, call it $n^w=[n^w_x,n^w_y,n^w_z]^\intercal$, in the world frame.  To convert the normal vector in the body frame, $n^r = [n^r_x,n^r_y,n^r_z]^\intercal$, we rotate by the robot's yaw: $n^r = R_\theta n^w$ (see Figure \ref{fig:slope_angle}), where $R_\theta$ is a basic rotation matrix by the angle $\theta$ about the world $z$ axis:
\begin{equation}
    R_\theta = \begin{bmatrix}
        \cos{p_\theta} && \sin{p_\theta} && 0\\
        -\sin{p_\theta} && \cos{p_\theta} && 0\\
        0 && 0 && 1
    \end{bmatrix}
\end{equation}
Let the robot pitch and roll vector $\omega$ be defined as $\omega = g(n^r)$, where:
\begin{align}
\omega = g(n^r) =
\begin{bmatrix}
    \mathrm{atan2}(n^r_x,n^r_z)\\
    -\mathrm{atan2}(n^r_y,n^r_z)
\end{bmatrix}
\end{align}
Creating a linearly-constrained problem requires a linear approximation of the constraint: 
\begin{equation}
    |\nabla_s\omega(s)\delta s + \omega(s)| <= \omega_{max}
\end{equation}
Conveniently, computing $\nabla_s\omega(s)$ reduces to finding gradients w.r.t position and yaw separately.  Let $\nabla_s\omega(s) = [\nabla_p\omega(s), \nabla_\theta\omega(s)]^\intercal$, then:
\begin{align}
    \nabla_p\omega(s) &= (\nabla_{n^r}g)(R_\theta)(\nabla_{p} n^w) \\
    \nabla_\theta\omega(s) &= (\nabla_{n^r}g)(\frac{d}{d\theta}R_\theta)(n^w)
\end{align}
where:
\begin{equation}
    \nabla_{n^r}g = \begin{bmatrix}
        \frac{n^r_z}{(n^r_x)^2+(n^r_z)^2} && 0 && \frac{-n^r_x}{(n^r_x)^2+(n^r_z)^2} \\
        0 && \frac{-n^r_z}{(n^r_y)^2+(n^r_z)^2} && \frac{n^r_y}{(n^r_y)^2+(n^r_z)^2}
    \end{bmatrix}
\end{equation}
and 
\begin{equation}
    \nabla_p n^w = \begin{bmatrix}
        \frac{\partial n^w_x}{\partial p_x} && \frac{\partial n^w_x}{\partial p_y} \\
        \frac{\partial n^w_y}{\partial p_x} && \frac{\partial n^w_y}{\partial p_y} \\
        \frac{\partial n^w_z}{\partial p_x} && \frac{\partial n^w_z}{\partial p_y} \\
    \end{bmatrix}
\end{equation}
The terms with the form $\frac{\partial n^w_x}{\partial p_x}$ amount to computing a second-order gradient of the elevation on the 2.5D map.  This can be done efficiently with numerical methods \cite{Fankhauser2016GridMapLibrary}.

\section{Converting non-linear MPC problem to a QP problem}
\label{appendix:qp}
Our MPC problem stated in Equations (\ref{eq:mpc_cost}-\ref{eq:mpc_state_constraints}) is non-linear.  In order to efficiently find a solution we linearize the problem about an initial solution, and solve iteratively, in a sequential quadratic programming (SQP) fashion \cite{nocedal2006numerical}.  Let $\{\hat{x}_k, \hat{u}_k\}_{k=0,\cdots,T}$ denote an initial solution.  Let $\{\delta x_k, \delta u_k\}_{k=0,\cdots,T}$ denote deviation from the initial solution.  We approximate (\ref{eq:mpc_cost}-\ref{eq:mpc_state_constraints}) by a problem with quadratic costs and linear constraints with respect to $\{\delta x, \delta u\}$:
\begin{align}
 \{\delta x^*, \delta u^*\} &= \argmin_{\delta x, \delta u}
    \sum_{k=0}^T \|\hat{x}_{k} + \delta x_k - x^*_{k}\|_{Q_k} \nonumber\\
    &\qquad \qquad \qquad + \lambda J(\hat{x}_k + \delta x_k, \hat{u}_k + \delta u_k)
    \label{eq:mpc_linear_start}\\
 s.t. \quad \forall k&\in[0,\cdots,T]:\nonumber
 \end{align}
 \begin{align}
    \hat{x}_{k+1} + \delta x_{k+1} &= f(\hat{x}_k, \hat{u}_k) + \nabla_x f\cdot\delta x_k + \nabla_u f\cdot\delta u_k \label{eq:mpc_linear_constraints_start}\\ 
    g(\hat{u_k}) &+ \nabla_u g\cdot\delta u_k \succ 0 \\ 
    h(m,\hat{x}_k) &+\nabla_x h\cdot\delta x_k \succ 0 
    \label{eq:mpc_linear_end}
\end{align}
where $J(\hat{x}_k + \delta x_k, \hat{u}_k + \delta u_k)$ can be approximated with a second-order Taylor approximation (for now, assume no dependence on controls):
\begin{align}
    J(\hat{x} + \delta x) \approx J(\hat{x}) + \nabla_x J\cdot\delta x + \delta x^\intercal H(J) \delta x
\end{align}
and $H(\cdot)$ denotes the Hessian.  The problem is now a quadratic program (QP) with quadratic costs and linear constraints.  To solve Equations (\ref{eq:mpc_linear_start}-\ref{eq:mpc_linear_end}), we introduce the solution vector variable $X$:
\begin{equation}
    X = 
    \begin{bmatrix}
        \delta x_0^\Tr & \cdots & \delta x_T^\Tr & \delta u_0^\Tr & \cdots & \delta u_T^\Tr\\
    \end{bmatrix}^\Tr
\end{equation}
We can then write Equations (\ref{eq:mpc_linear_start}-\ref{eq:mpc_linear_end}) in the form:
\begin{align}
    \text{minimize}\quad & \frac{1}{2}X^\Tr P X + q^\Tr X\\
    \text{subject to}\quad & l\le AX \le u
\end{align}
where $P$ is a positive semi-definite weight matrix, $q$
is a vector to define the first order term in the objective function, $A$ defines inequality constraints and $l$ and $u$ provide their lower and upper limit. 

\section{Linesearch Algorithm for SQP solution refinement}
\label{appendix:linesearch}

The solution to the SQP problem returns an optimized control sequence $\{u_k^*\}_{k=0:T}$.  We then use a linesearch routine to find an appropriate correction coefficient $\gamma$, using Algorithm \ref{alg:linesearch}. The resulting correction coefficient is carried over into the next path-planning loop.
\vspace{5mm}

\begin{algorithm}
    \caption{Linesearch Algorithm}
    \begin{algorithmic}[1]
        \renewcommand{\algorithmicrequire}{\textbf{Input:}}
        \renewcommand{\algorithmicensure}{\textbf{Output:}}
        \REQUIRE candidate control sequence $\{u^\mathrm{c}_k\}_{k=0:T}$, QP solution $\{\delta u^*_k\}_{k=0:T}$
        \ENSURE  correction coefficient $\gamma$
        \\ \textit{Initialization}
        \STATE initialize $\gamma$ by default value or last-used value
        \STATE $[c, o]=$getCostAndObstacles($\{u^\mathrm{c}_k\}_{k=0:T}$)
        \\ \textit{Linesearch Loop}
        \FOR {$i = 0$ to $\text{max\_iteration}$}
        \FOR {$k = 0$ to $T$}
        \STATE $u^{\mathrm{c}(i)}_k=u^\mathrm{c}_k + \gamma \delta u^*_k$
        \ENDFOR
        \STATE $[c^{(i)}, o^{(i)}]=$getCostAndObstacles($\{u^{\mathrm{c}(i)}_k\}_{k=0:T}$)
        \IF {($c^{(i)} \le c$ and $o^{(i)} \le o$)}
        \STATE $\gamma=\min(2\gamma,\gamma_{max})$
        \BREAK
        \ELSE
        \STATE $\gamma=\max(\gamma/2,\gamma_{min})$
        \ENDIF
        \ENDFOR
        \RETURN $\gamma$ 
    \end{algorithmic} 
    \label{alg:linesearch}
\end{algorithm}

\end{document}